\documentclass[10pt,twocolumn,letterpaper]{article}

\usepackage[pagenumbers]{cvpr} 

\usepackage{times}
\usepackage{epsfig}
\usepackage{graphicx}
\usepackage{amsmath}
\usepackage{amssymb}
\usepackage{booktabs}
\usepackage{multirow}
\usepackage{leftidx}
\usepackage{bbm}
\usepackage{floatrow}

\usepackage{algorithmic}
\usepackage{textcomp}
\usepackage{siunitx}
\usepackage{tablefootnote}
\usepackage{footnote}
\usepackage{caption}
\usepackage{makecell}
\usepackage{pifont}
\newcommand{\cmark}{\ding{51}}%
\newcommand{\xmark}{\ding{55}}%
\usepackage[pagebackref,breaklinks,colorlinks]{hyperref}
\hypersetup{colorlinks,linkcolor={blue},citecolor={blue},urlcolor={blue}}  

\usepackage[capitalize]{cleveref}
\crefname{section}{Sec.}{Secs.}
\Crefname{section}{Section}{Sections}
\Crefname{table}{Table}{Tables}
\crefname{table}{Tab.}{Tabs.}
\def\cvprPaperID{***} 
\def\confName{CVPR}
\def\confYear{2023}

\usepackage{xcolor}
\usepackage{enumitem}
\usepackage[accsupp]{axessibility}  

\newcolumntype{P}[1]{>{\centering\arraybackslash}p{#1}}
\newcolumntype{M}[1]{>{\centering\arraybackslash}m{#1}}

\newcommand{\net}{\phi}
\newcommand{\inpt}{\mathbf{x}}
\newcommand{\TrSet}{\mathit{S}}
\newcommand{\TeSet}{\mathit{T}}
\newcommand{\param}{\mathbf{\theta}}
\newcommand{\optParam}{\hat{\param}}

\newcommand{\mean}{\mu}
\newcommand{\var}{\sigma^2}
\newcommand{\trmean}{\hat{\mu}}
\newcommand{\trvar}{\hat{\sigma}^2}

\newcommand{\layer}{l}
\newcommand{\Layers}{L}
\newcommand{\voxel}{v}
\newcommand{\Voxels}{V}
\newcommand{\nchannels}{c_\layer}
\newcommand{\nframes}{t_\layer}
\newcommand{\height}{h_\layer}
\newcommand{\width}{w_\layer}
\newcommand{\voxelrange}{[1,\nframes]\times[1,\height]\times[1,\width]}

\newcommand{\mycomment}[1]{}

\newcommand{\myparagraph}[1]{\vspace{2pt}\noindent{\bf #1}}

\newcommand{\OurMethod}{ViTTA\xspace}

\newif\ifdraft
\draftfalse
\drafttrue 

\definecolor{orange}{rgb}{1,0.5,0}
\definecolor{violet}{RGB}{70,0,170}
\definecolor{magenta}{RGB}{170,0,170}
\definecolor{dgreen}{RGB}{0,150,0}

\usepackage[normalem]{ulem}

\ifdraft
 \newcommand{\Wei}[1]{{\color{blue}{\bf Wei: #1}}}
 
 \newcommand{\MK}[1]{{\color{magenta}{\bf MK: #1}}}
 
\else
 \newcommand{\Wei}[1]{}
 
 \newcommand{\MK}[1]{}
 
\fi

\makeatletter
\def\blfootnote{\gdef\@thefnmark{}\@footnotetext}
\makeatother

\begin{document}
\title{Video Test-Time Adaptation for Action Recognition}


\author{
Wei Lin$^{\dagger\ast 1,2}$ \and
Muhammad Jehanzeb Mirza$^{\ast1,3}$ \and 
Mateusz Kozinski$^1$ \and 
Horst Possegger$^1$ \and 
Hilde Kuehne$^{4,5}$ \and
Horst Bischof$^{1,3}$\and\\
$^1$Institute for Computer Graphics and Vision, Graz University of Technology, Austria\\
$^2$Christian Doppler Laboratory for Semantic 3D Computer Vision\\
$^3$Christian Doppler Laboratory for Embedded Machine Learning\\
$^4$Goethe University Frankfurt, Germany\\
$^5$MIT-IBM Watson AI Lab
}

\maketitle
\blfootnote{
$\ast$ Equally contributing authors.\\ 
{
\indent \indent$\dagger$ Correspondence: \tt\small{wei.lin@icg.tugraz.at}}
}

\begin{abstract}
Although action recognition systems can achieve top performance when evaluated on in-distribution test points, they are vulnerable to unanticipated distribution shifts in test data. However, test-time adaptation of video action recognition models against common distribution shifts has so far not been demonstrated. We propose to address this problem with an approach tailored to spatio-temporal models that is capable of adaptation on a single video sample at a step. It consists in a feature distribution alignment technique that aligns online estimates of test set statistics towards the training statistics. We further enforce prediction consistency over temporally augmented views of the same test video sample. Evaluations on three benchmark action recognition datasets show that our proposed technique is architecture-agnostic and able to significantly boost the performance on both, the state of the art convolutional architecture TANet and the Video Swin Transformer. Our proposed method demonstrates a substantial performance gain over existing test-time adaptation approaches in both evaluations of a single distribution shift and the challenging case of random distribution shifts. Code will be available at \url{https://github.com/wlin-at/ViTTA}.
\end{abstract}

\section{Introduction}
State-of-the-art neural architectures~\cite{feichtenhofer2020x3d,wang2021action,yang2020temporal,xiang2022spatiotemporal,yang2022recurring,truong2022direcformer} are very effective in action recognition, but recent work shows they are not robust to shifts in the distribution of the test data~\cite{yi2021benchmarking,schiappa2022large}.
Unfortunately, in practical scenarios, such distribution shifts are very difficult to avoid, or account for.
For example, cameras used for recognizing motorized or pedestrian traffic events may register rare weather conditions, like a hailstorm,
and sports action recognition systems can be affected by perturbations generated by spectators at sports arenas, such as the smoke of flares.
Shifts in the data distribution can also result from inconspicuous changes in the video processing setup, for instance, a change of the algorithm used to compress the video feed.
\begin{figure}[t]
\includegraphics[width=\columnwidth]{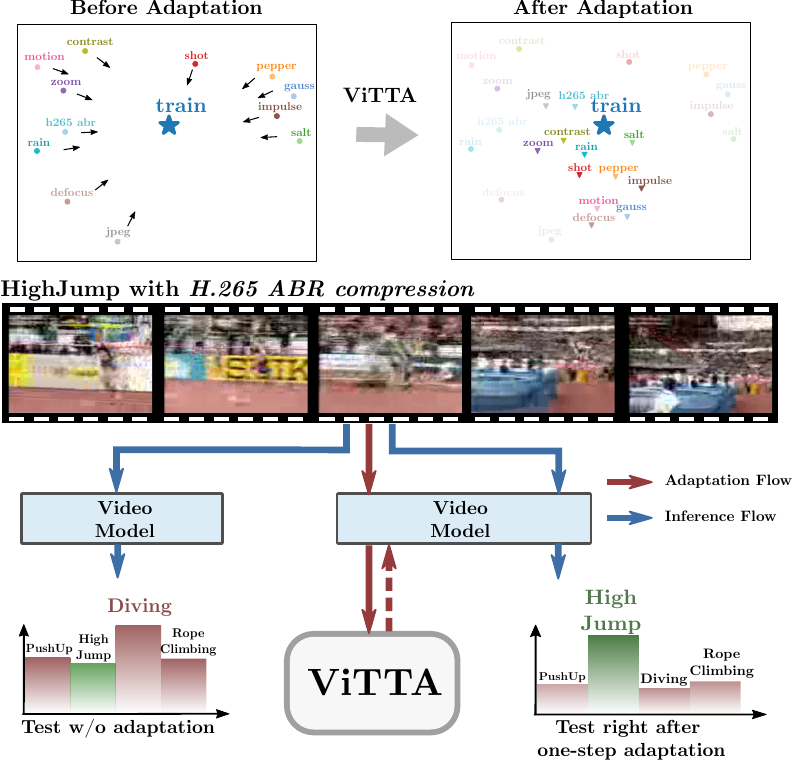}
\caption{
\label{fig:teaser}
\iftrue
Our architecture-agnostic \OurMethod enables action recognition models to overcome severe video corruptions.
In a fully online manner, \ie~processing each test video only once, we perform test-time adaptation.
In particular, we align the statistics of the (corrupted) test data towards the (clean) training data such that they are better aligned (top: t-SNE~\cite{van2008visualizing} of mean features from final feature extractor layer for clean training data and for 12 differently corrupted test datasets). 
This results in a significant performance improvement for action recognition.
\fi
}
\end{figure}

In image classification, distribution shift can be mitigated by Test-Time-Adaptation (TTA)~\cite{wang2021tent,liang2020shot,iwasawa2021test,schneider2020norm,sun2020ttt,liu2021tttpp,mirza2022dua,kojima2022cfa}, which uses the unlabeled test data to adapt the model to the change in data distribution.
However, methods developed for image classification are not well suited for action recognition.
Most action recognition applications require running memory- and computation-hungry temporal models online, with minimal delay, and under tight hardware constraints.
Moreover, videos are more vulnerable to distribution shifts than images~\cite{yi2021benchmarking,schiappa2022large,azimi2022self}. Some examples of such distribution shifts are given in Fig.~\ref{fig:teaser}. 
Due to limited exposure times, video frames are likely to feature higher variations of noise level, provoked by illumination changes. 
They are more affected by motion blur, which varies with the speed of motion observed in the scene.
They also feature stronger compression artifacts, which change with the compression ratio, often dynamically adjusted to the available bandwidth.
Our experiments show that existing TTA algorithms, developed for image classification, do not cope with these challenges well, yielding marginal improvement over networks used on corrupted data without any adaptation.

Our goal is to propose an effective method for online test-time-adaptation of action recognition models.
Operating online and with small latency can require drastically constraining the batch size, especially when the hardware resources are limited and the employed model is large.
We therefore focus on the scenario in which test samples are processed individually, one at a time.
To ease the integration of our method in existing systems, we require it to be capable of adapting pretrained networks, both convolutional and transformer-based, without the need to retrain them.

In the work, we propose the first video test-time adaptation approach - ViTTA. To address the above requirements, we turn to feature alignment~\cite{sun2016coral,zellinger2017central,schneider2020norm,mirza2022dua,kojima2022cfa}, a common TTA method that aligns distributions of features computed for test and training data by minimizing discrepancy between their statistics.
Feature alignment does not require any modifications to the training procedure and is architecture-agnostic.
However, existing feature alignment methods are not well suited for online adaptation, because they require relatively large test batches to accurately estimate the statistics.
We address this by employing the exponential moving average to estimate test feature statistics online.
This enables us to perform the alignment by processing one video sample at a time, at a low computational and memory cost.
Additionally, we show that even though the temporal dimension of video data poses challenges, it also has a silver lining.
We leverage this by creating augmented views of the input videos via temporally resampling frames in the video. This has two benefits: First, multiple augmented views lead to more accurate statistics of the overall video content. Second, it allows us to enforce prediction consistency across the views, making the adaptation more effective.

Our extensive evaluations on three most popular action recognition benchmarks demonstrate that \OurMethod boosts the performance of both TANet~\cite{liu2021tam}, the state-of-the-art convolutional architecture, and the Video Swin Transformer~\cite{liu2022video}, and outperforms the existing TTA methods proposed for image data by a significant margin. \OurMethod performs favorably in both, evaluations of single distribution shift and the challenging case of random distribution shift. 

\OurMethod also has a high practical valor.
It is fully online and applicable in use cases requiring minimal delays. It does not require collection or storage of test video dataset, which is significant in terms of data privacy protection, especially in processing confidential user videos.
\OurMethod can be seamlessly incorporated in systems already in operation, as it has no requirement of re-training existing networks. Therefore it can harness state-of-the-art video architectures. 

Our contributions can be summarized as follows:\\
\begin{itemize}[nosep] 
  \item We benchmark existing TTA methods in online adaptation of action recognition models to distribution shifts in test data, on three most popular action recognition datasets, UCF101~\cite{soomro2012ucf101}, Something-something v2~\cite{goyal2017something} and Kinetics 400~\cite{kay2017kinetics} \\
  \item We adapt the feature alignment approach to online action recognition, generating a substantial performance gain over existing techniques.\\
  \item We propose a novel, video-specific adaptation technique (\OurMethod) that enforces consistency of predictions for temporally re-sampled frame sequences and show that it contributes to adaptation efficacy. \\
\end{itemize}

 \section{Related Work}

\textbf{Action Recognition} is addressed mainly with CNN-based and transformer-based architectures. CNN-based architectures typically use 3D convolutions, such as C3D~\cite{tran2015learning}, I3D~\cite{carreira2017quo}, Slowfast~\cite{feichtenhofer2019slowfast} and X3D~\cite{feichtenhofer2020x3d}. Recent works also deploy 2D convolution with temporal modules~\cite{wang2016temporal,qiu2017learning,lin2019tsm,liu2020teinet,li2020tea,wang2021tdn,liu2021tam} to reduce the computational overhead. TEINet~\cite{liu2020teinet} learns the temporal features by decoupling the modeling of channel correlation and temporal interaction for efficient temporal modeling. TANet~\cite{liu2021tam} is a 2D CNN with integrated temporal adaptive modules which generates temporal kernels from its own feature maps, achieving state-of-the-art performance among convolutional competitors. Transformer-based models have also been applied for video recognition~\cite{bertasius2021space,arnab2021vivit,liu2022video,zhang2021token,fan2021multiscale,neimark2021video,patrick2021keeping}. ViViT~\cite{arnab2021vivit} adds several temporal transformer encoders on the top of spatial encoders.
Video swin transformer~\cite{liu2022video} uses spatio-temporal local windows to compute the self-attention. In this work, we evaluate our adaptation method on TANet and Video Swin Transformer. 

\textbf{Robustness of Video Models} for action recognition against common corruptions has recently been analyzed~\cite{yi2021benchmarking,schiappa2022large}. Yi \etal ~\cite{yi2021benchmarking} and Schiappa \etal~\cite{schiappa2022large} benchmark robustness of common convolutional- and transformer-based spatio-temporal architectures, against several corruptions in video acquisition and video processing. In this work, we perform evaluations on 12 corruptions proposed in these two benchmark works. These corruptions cover various types of noise and digital errors, blur effects of cameras, weather conditions, as well as quality degradation in image and video compression.


\textbf{Test-Time Adaptation} tackles the adaptation to unknown distribution shifts encountered at test-time in an unsupervised manner. It has recently gained increasing attention in the image domain~\cite{sun2020ttt,liu2021tttpp,gandelsman2022test,wang2021tent,zhang2021memo,liang2020shot,boudiaf2022parameter,iwasawa2021t3a,mirza2022dua,schneider2020norm}. These approaches can be divided into two distinct groups:
The first group modifies the training procedure and employs a self-supervised auxiliary task to adapt to distribution shifts at test-time. 
Sun~\etal~\cite{sun2020ttt} train the network jointly for self-supervised rotation prediction~\cite{gidaris2018unsupervised} and the main task of image classification. 
At test-time they adapt to the out-of-distribution test data by updating the encoder through the gradients obtained from the auxiliary task of rotation prediction on the test samples. 
TTT++~\cite{liu2021tttpp} propose self-supervised contrastive learning~\cite{chen2020simple} as an auxiliary objective for adaptation and also aligns the source and target domain feature responses. 
Recently, Gandelsman~\etal~\cite{gandelsman2022test} use the self-supervised reconstruction task through masked autoencoders~\cite{he2022masked} for test-time adaptation. 

The other group of methods, more closely related to our work, propose to adapt off-the-shelf pre-trained networks without altering the training. 
These methods typically employ post-hoc regularization. 
TENT~\cite{wang2021tent} adapts a pre-trained network at test-time by minimizing the entropy from the output softmax distribution. 
Similarly, MEMO~\cite{zhang2021memo} proposes to adapt the network at test-time by minimizing the entropy of the marginal output distribution across augmentations. 
SHOT~\cite{liang2020shot} also uses entropy minimization and adds information maximization regularization at test-time. 
On the other hand, some methods do not update the parameters for the network and instead propose gradient-free approaches for test-time adaptation. 
For example, LAME~\cite{boudiaf2022parameter} only adapts the outputs of the network by using Laplacian regularization and guarantees convergence through a concave-convex optimization procedure. T3A~\cite{iwasawa2021t3a} generates pseudo-prototypes from the test samples and replaces the classifier learned on the training set. DUA~\cite{mirza2022dua} and NORM~\cite{schneider2020norm} only update the statistics of the batch normalization layer for adaptation at test-time. 

Despite the intensive development in the image domain, test-time adaptation on video action recognition models, to the best of our knowledge, has not been demonstrated so far. 
In this work, we propose a video-tailored adaptation method - \OurMethod that adapts spatio-temporal models against common distribution shifts on video sequences. \OurMethod consists in a feature distribution alignment technique that aligns training statistics with the online estimates of test set statistics. It is model-agnostic and can adapt to both convolutional and transformer-based network without the need to re-train them. 
We compare \OurMethod to existing TTA approaches that do not require to alter the training of source model, and adapt to off-the-shelf pretrained-models. 
\section{Video Test-Time Adaptation (ViTTA)}

\begin{figure*}
\includegraphics[width=0.9\columnwidth]{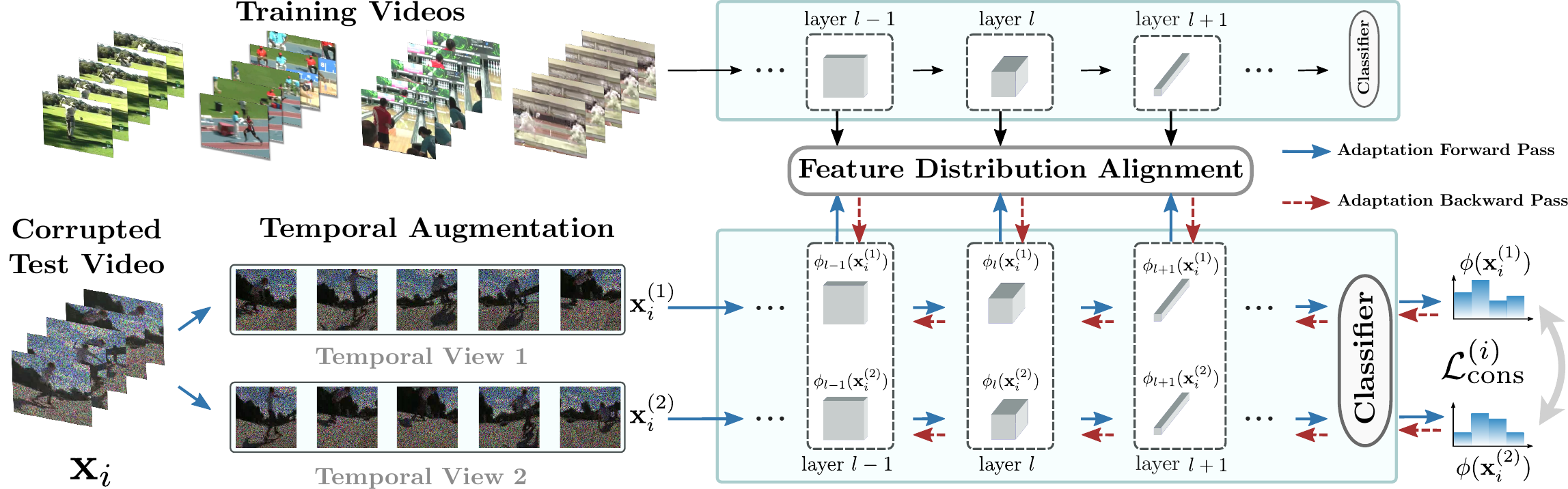}
\caption{
\label{fig:pipeline}
Pipeline of \OurMethod. The online adaptation is applied on videos that are received sequentially and here we show the adaptation process of iteration $i$. We first compute the online estimates of the test statistics by 1) sampling two temporally augmented views from the test video, and computing the statistics on multi-layer features maps across the two views, 2) then performing exponential moving averages of statistics among iterations. Afterwards, we perform feature distribution alignment by minimizing the discrepancy between the pre-computed training statistics and the online estimates of test statistics. Furthermore, we enforce prediction consistency over temporally augmented views for performance boost. 
}
 
\end{figure*}
We are given a multi-layer neural network $\net$ trained in action recognition on a training set of video sequences $\TrSet$, and its optimal parameter vector $\optParam$, resulted from this training.
At test time, the network is exposed to unlabeled videos from the test set $\TeSet$, that may be distributed differently than the data in $\TrSet$.
Our goal is to adapt $\net$ to this distribution shift to maximize its performance on the test videos. 
The pipeline of our method - ViTTA is shown in Fig.~\ref{fig:pipeline}.

\subsection{Feature distribution alignment}
We perform the adaptation by aligning the distribution of feature maps computed for the training and test videos.
Following recent work on TTA~\cite{schneider2020norm,mirza2022dua}, to align the distributions, we equalize means and variances of the feature maps.
We denote the feature map of the $\layer$-th layer of the network $\net$, computed for a video $\inpt$, by $\net_{\layer}(\inpt; \param)$, where $\param$ is the parameter vector used for the computation. The feature map is a tensor of size $(\nchannels,\nframes,\height,\width)$, where $\nchannels$ denotes the number of channels in the $\layer$-th layer and $\nframes$, $\height$, and $\width$ are its temporal and spatial dimensions. 
We denote the spatio-temporal range of the feature map by $\Voxels=\voxelrange$, and a $c_\layer$-element feature vector at voxel $\voxel \in \Voxels$ by $\net_{\layer}(\inpt;\param)[\voxel]$.
The mean vector of the $\layer$-th layer features for dataset $D$ can be computed as the sample expectation
\begin{equation}
    \mean_{\layer} (D;\param) = 
        \mathbb{E}_{\substack{\inpt \in D\\ \voxel \in \Voxels}} 
            \big[ \net_{\layer} (\inpt;\param)[\voxel] \big] ,
\end{equation}
and the vector of variances of the $\layer$-th layer features can be obtained as
\begin{equation}
    \var_{\layer} (D;\param) = 
        \mathbb{E}_{\substack{\inpt \in D\\ \voxel \in \Voxels}} 
            \left[ \big( \net_{\layer} (\inpt;\param)[\voxel] - \mean_{\layer} (D;\param) \big)^2 \right] .
\end{equation}
To declutter the equations, we shorten the notation of the training statistics to $\trmean_\layer=\mean_{\layer}(\TrSet;\optParam)$ and $\trvar_\layer=\var_\layer(\TrSet;\optParam)$.
In our experiments, we pre-compute them on the training data. When this data is no longer available, they can be substituted with statistics of another, unlabeled dataset, that is known to be generated from a similar distribution. 
In Sec.~\ref{sec:statistics_norm_layers}, we also show that 
for a network with batch normalization layers, running means and variances accumulated in these layers can be used instead of the statistics computed for the training set, with a small performance penalty.

Our overarching approach is to iteratively update the parameter vector $\param$ in order to align the test statistics of selected layers to the statistics computed for the training data.
This can be formalized as minimizing the alignment objective
\begin{equation} \label{eq:offline_objective}
\mathcal{L}_\mathrm{align}(\param)=\sum_{\layer \in \Layers} \lvert \mean_{\layer}(\TeSet;\param) - \trmean_\layer \rvert \\
                                                           + \lvert  \var_{\layer}(\TeSet;\param) -  \trvar_\layer \rvert
\end{equation}
with respect to the parameter vector $\param$,
where $L$ is the set of layers to be aligned, $|\cdot|$ denotes the vector $l_1$ norm, and we recall that $\TeSet$ denotes the test set.
Our approach of \emph{training} the network to align the distributions is different in spirit from the TTA techniques based on feature alignment~\cite{nado2020evaluating,schneider2020norm,mirza2022dua}, which only adjust the running statistics accumulated during training in the normalization layers
and therefore do not truly learn at test time.
The fact that we update the entire parameter vector differentiates our method from the existing algorithms that only update parameters of affine transformation layers~\cite{wang2021tent,kojima2022cfa}, and gives it more flexibility during adaptation.
Even though our method adjusts the full parameter vector, our experiments on continued adaptation (in Sec.~\ref{sec:continuous_adapt}), show that our methods adapts fast to periodic changes of distribution shift.
When the distribution shift is removed from the stream of test data, the network quickly restores its original performance.
We also found that
the best performance was achieved by aligning the distributions of the features output by the last two out of four blocks for both TANet~\cite{liu2021tam} and Video Swin Transformer~\cite{liu2022video}. We therefore set $\Layers$ to contain layers in these two blocks. The respective ablation study can be found in Sec.~\ref{sec:num_views_pred_consis}.

\subsection{Online adaptation}
Optimizing the objective in Eq.~\eqref{eq:offline_objective} requires iteratively estimating statistics of the test set. 
This is infeasible in an online video recognition system, typically required to process a stream of data with minimal delay.
We therefore adapt the feature alignment approach to an online scenario. 
We assume the test data is revealed to the adaptation algorithm in a sequence of videos, denoted as $\inpt_i$, where $i$ is the index of the test videos.
We perform one adaptation step for each element of the sequence.
Feature statistics computed on a single test sample do not represent feature distribution over the entire test set,
so we cannot only rely on them when aligning the distributions.
We therefore approximate test set statistics by exponential moving averages of statistics computed on consecutive test videos
and use them for the alignment.
We define the mean estimate in iteration $i$ as
\begin{equation} \label{eq:runnin_mean}
    {\mean_{\layer}}^{(i)}(\param)=\alpha\cdot \mean_{\layer}(\inpt_i;\param) + (1-\alpha)\cdot {\mean_{\layer}}^{(i-1)}(\param),
\end{equation}
where $1-\alpha$ is the momentum that is set to a common choice of $0.9$ ($\alpha=0.1$).
Similarly, we define the $i$-th variance estimate as
\begin{equation} \label{eq:runnin_var}
    {\var_{\layer}}^{(i)}(\param)=\alpha\cdot \var_{\layer}(\inpt_i;\param) + (1-\alpha)\cdot {\var_{\layer}}^{(i-1)}(\param).
\end{equation}
To suit online adaptation, in the $i$-th alignment iteration, the objective in Eq.~\eqref{eq:offline_objective} is approximated by
\begin{equation}
\mathcal{L}^{(i)}_\mathrm{align}(\param)=\sum_{\layer \in \Layers} \lvert {\mean_{\layer}}^{(i)}(\param) - \trmean_\layer \rvert \\
                                                                 + \lvert  {\var_{\layer}}^{(i)}(\param) -  \trvar_\layer \rvert.
\end{equation}
This approach simultaneously decreases the variance of the estimates and lets the network continuously adapt to changing distribution of test data. 

\subsection{Temporal augmentation}
To further increase the efficacy of the method, we benefit from the temporal nature of the data and create $M$ resampled views of the same video. We denote the temporally augmented views of the input video by $\inpt^{(m)}_i$, for $1\le m \le M$. 
We compute the mean and variance vector of video $\inpt_i$ over the $M$ views, to improve the accuracy of statistics on a single video: 
\begin{equation}
    \mean_{\layer} (\inpt_i;\param) = 
        \mathbb{E}_{\substack{m \in M\\ \voxel \in \Voxels}} 
            \big[ \net_{\layer} (\inpt^{(m)}_i;\param)[\voxel] \big] ,
\end{equation}
\begin{equation}
    \var_{\layer} (\inpt_i;\param) = 
        \mathbb{E}_{\substack{m \in M\\ \voxel \in \Voxels}} 
            \left[ \big( \net_{\layer} (\inpt^{(m)}_i;\param)[\voxel] - \mean_{\layer} (\inpt_i;\param) \big)^2 \right] .
\end{equation}
We recall that $\mean_{\layer} (\inpt_i;\param)$ and $\var_{\layer} (\inpt_i;\param)$ are used in Eq.~\eqref{eq:runnin_mean} and \eqref{eq:runnin_var} for computing mean and variance estimate in iteration $i$.

Furthermore, we enforce consistency of the corresponding predictions among the $M$ views. 
We establish a pseudo-label by averaging the class probabilities predicted by the network for the input views $y(\inpt)=\frac{1}{M} \sum_{m=1}^{M} \net(\inpt^{(m)}_i;\param)$, 
 and define the consistency objective in iteration $i$ as
\begin{equation}
\mathcal{L}^{(i)}_\mathrm{cons} (\param)=  \sum_{m=1}^{M} \lvert \net (\inpt^{(m)}_i;\param) - y(\inpt) \rvert.
\end{equation}
In the $i$-th alignment iteration, we update the network parameter by following the gradient of 
\begin{equation}
\min_{\param} \mathcal{L}^{(i)}_\mathrm{align}(\param) + \lambda\cdot \mathcal{L}^{(i)}_\mathrm{cons} (\param), 
\end{equation}
where $\lambda$ is the coefficient that we set to 0.1. In the ablation study, we show that setting $M=2$ is enough to deliver a significant performance boost (Sec.~\ref{sec:num_views_pred_consis}), and that uniform equidistant resampling of the input videos yields the best results (Sec.~\ref{sec:temp_samp_strategies}).


\section{Experiments}
\label{sec:experiments}
\subsection{A Benchmark for Video Test-Time Adaptation}
\subsubsection{Video datasets.} We conduct experiments on three action recognition benchmark datasets: UCF101~\cite{soomro2012ucf101}, Something-something v2 (SSv2)~\cite{goyal2017something} and Kinetics 400 (K400)~\cite{kay2017kinetics}.
The UCF101 dataset contains 13320 videos collected from YouTube with 101 action classes. We evaluate on split 1, which consists of 9537 training videos and 3783 validation videos. SSv2 is a large-scale action dataset with 168K training videos and 24K validation videos, comprised of 174 classes. K400 is the most popular benchmark for action recognition tasks, containing around 240K training videos and 20K validation videos in 400 classes. 

\subsubsection{Corruptions.} We evaluate on 12 corruption types proposed in \cite{schiappa2022large,yi2021benchmarking} that benchmark robustness of spatio-temporal models.
These 12 corruptions are: \textit{Gaussian noise}, \textit{pepper noise}, \textit{salt noise}, \textit{shot noise}, \textit{zoom blur}, \textit{impulse noise}, \textit{defocus blur}, \textit{motion blur}, \textit{jpeg compression}, \textit{contrast}, \textit{rain}, \textit{H.265 ABR compression}.
They cover various types of noise and digital errors, blur effects of cameras, weather conditions, as well as quality degradation in image and video compression. 

Following \cite{schiappa2022large,yi2021benchmarking}, we evaluate on the validation sets of the three datasets. We use the implementation from these two benchmark papers for corruption generation. As the robustness analysis indicates approximate linear correlation between severity level and the performance drop, 
We evaluate on the corruptions of the most severe case at level 5. 

\subsection{Implementation Details}
We evaluate our approach on two model architectures: TANet~\cite{liu2021tam} based on ResNet50~\cite{he2016deep}, and Video Swin Transformer~\cite{liu2022video} based on Swin-B~\cite{liu2021swin}. 
TANet is the state-of-the-art convolutional network for action recognition and Video Swin Transformer is adapted from Swin Transformer~\cite{liu2021swin}. 
More model specifics are given in supplementary.

We perform distribution alignment of features from the normalization layers in the last two blocks. 
We set the batch size to 1 for evaluation on all datasets. We adapt to each video only once. Following common practice in online test-time adaptation~\cite{sun2020ttt,wang2022continual,boudiaf2022parameter}, we perform inference right after adapting to a sample and report the accumulated accuracy for all samples. On TANet, we set the learning rate to $5e-5$ for UCF and $1e-5$ for SSv2 and K400. On Video Swin Transformer, learning rate is set to $1e-5$ for all datasets.
For the temporal augmentation, we perform uniform equidistant sampling and random spatial cropping. 




\subsection{Comparison to State-of-the-Art}
We evaluate our adaptation algorithm against the following baselines that adapt to off-the-shelf pretrained-models without altering the training conditions. 
\textit{Source-Only} denotes generating the predictions directly with the model trained on the training data, without adaptation.
NORM~\cite{schneider2020norm} consists in adapting statistics of batch normalization layers to test data. 
DUA~\cite{mirza2022dua} adapts the batch normalization layers online.
TENT~\cite{wang2021tent} learns affine transformations of feature maps by minimizing the entropy of test predictions.
SHOT (online)~\cite{liang2020shot} maximizes the entropy of batch-wise distribution of predicted classes, while minimizing the entropy of individual predictions.
T3A~\cite{iwasawa2021test} builds pseudo-prototypes from test data.

\begin{table}[!tb]
\setlength\tabcolsep{4pt}
    \centering
    \scriptsize
    \begin{tabular}{lcccccc}
    \toprule
    \multicolumn{7}{c}{Batch size 1} \\
    \toprule
    Model & \multicolumn{3}{c}{TANet} & \multicolumn{3}{c}{Swin} \\
    \cmidrule(r){2-4}\cmidrule{5-7}
    Dataset & UCF101 & SSv2 & K400 &  UCF101 &  SSv2 & K400\\
    \midrule
    clean & 96.67 & 59.98 & 71.64  & 97.30  & 66.36 & 75.32 \\
    source & 51.35 & 24.31 & 37.16 & 78.48 & 42.18 & 47.17 \\
    \midrule
    NORM & 51.59 & 17.21 & 31.40 & - & - & - \\
    DUA & \underline{55.34} & 16.20 & 31.88 & - & - & -  \\
    TENT & 51.58 & 17.29 & 31.43 & \underline{81.19}  & 36.05 & 45.18 \\
    SHOT & 51.20 & 14.54 & 25.17 & 68.51 & 21.32 & 29.22 \\
    T3A & 54.17 & \underline{24.42} & \underline{37.59} & 80.66 & \underline{42.41} & \underline{48.20} \\
    \midrule
    \OurMethod & \textbf{78.20} & \textbf{37.97} & \textbf{48.69} & \textbf{84.63} & \textbf{48.52} & \textbf{52.11}\\ 
    \bottomrule
    \label{tab:main-batchsize1}
    \end{tabular}

    \scriptsize
    \begin{tabular}{lcccccc}
    \toprule
    \multicolumn{7}{c}{Batch size 8} \\
    \toprule
    Model & \multicolumn{3}{c}{TANet} & \multicolumn{3}{c}{Swin} \\
    \cmidrule(r){2-4}\cmidrule{5-7}
    Dataset & UCF101 & SSv2 & K400 &  UCF101 &  SSv2 & K400\\
    \midrule
    clean & 96.67 & 59.98 & 71.64  & 97.30  & 66.36 & 75.32 \\
    source & 51.35 & 24.31 & 37.16 & 78.48 & 42.18 & 47.17 \\
    \midrule
    NORM & 65.77 & 28.18 & 39.32 & - & - & - \\
    TENT & \underline{72.92} & \underline{31.57} & \underline{41.13} & \underline{82.35}  & \underline{42.84} & 47.84 \\
    SHOT & 65.54 & 27.47 & 37.43 & 78.42 & 42.55 & 47.98 \\
    T3A & 54.17 &  24.45 & 37.59 & 80.68 & 42.41 & \underline{48.20} \\
    \midrule
    \OurMethod & \textbf{78.33} & \textbf{38.07} & \textbf{48.94} & \textbf{84.74} & \textbf{49.66} & \textbf{54.55}\\ 
    \bottomrule
    \label{tab:main-batchsize8}
    \end{tabular}
\caption{
Mean Top-1 Classification Accuracy (\%) for batch size 1 and 8 over all corruption types on UCF101, SSv2 and K400. We use the convolution-based TANet and a transformer-based Video Swin Transformer to evaluate our ViTTA and all other baselines. For a fair comparison with baselines, we provide results with different batch sizes, as some baselines require larger batch sizes for optimal performance. \textit{DUA} is only evaluated with batch size of 1, following the setting in the original work~\cite{mirza2022dua}. \emph{Clean} refers to the performance of the model on the original validation set of the respective datasets. \emph{Source} is the average performance of the source model over all corruption types without adaptation. Highest accuracy is shown in bold, while second best is underlined.
}
\vspace{-4mm}
\label{tab:main-batchsize-1_8}
\end{table}

\begin{table*}[!tb]
\scriptsize
\centering
\begin{tabular}{lM{0.75cm}M{0.75cm}M{0.75cm}M{0.75cm}M{0.75cm}M{0.75cm}M{0.75cm}M{0.75cm}M{0.75cm}M{0.75cm}M{0.75cm}M{0.75cm}M{0.75cm}}
\toprule
corruptions & gauss & pepper & salt & shot & zoom  & impulse &  defocus & motion &  jpeg & contrast & rain & h265.abr & avg\\
\midrule
Source-Only & 17.92 & 23.66 & 7.85 &  72.48 & 76.04 & 17.16 & 37.51 & 54.51 & 83.40 & 62.68 & 81.44 & 81.58 & 51.35 \\
\midrule
NORM & 45.23 & 42.43 & 27.91 & 86.25  & \underline{84.43} & 46.31 & 54.32 & 64.19 & 89.19 & 75.26 & 90.43 & 83.27 & 65.77\\
DUA & 36.61 & 33.97 & 22.39 & 80.25 & 77.13 & 36.72 & 44.89 & 55.67 & 85.12 & 30.58 & 82.66 & 78.14 & 55.34 \\
TENT & \underline{58.34} & \underline{53.34} & \underline{35.77} & \underline{89.61} & \textbf{87.68} & \underline{59.08} & \underline{64.92} & \underline{75.59} & \underline{90.99} & \underline{82.53} & \underline{92.12} & \textbf{85.09} & \underline{72.92} \\
SHOT & 46.10 & 43.33 & 29.50 & 85.51  & 82.95 & 47.53 & 53.77 & 63.37 & 88.69 & 73.30 & 89.82 & 82.66 &  65.54\\
T3A & 19.35 & 26.57 & 8.83 & 77.19  & 79.38 & 18.64 & 40.68 & 58.61 & 86.12 & 67.22 & 84.0 & 83.45 & 54.17\\
\midrule
\OurMethod & \textbf{71.37} & \textbf{64.55} & \textbf{45.84} & \textbf{91.44}  & \textbf{87.68} & \textbf{71.90} & \textbf{70.76} & \textbf{80.32} & \textbf{91.70} & \textbf{86.78} & \textbf{93.07} & \underline{84.56} & \textbf{78.33}\\

\bottomrule
\end{tabular}
\vspace{-2mm}
\caption{
\label{tab:resultsCorruptionWise_tanet_ucf}
Top-1 Classification Accuracy (\%) for all corruptions in the UCF101 dataset, while using the TANet backbone.
\textit{DUA} is evaluated with batch size of 1, following the setting in the original work~\cite{mirza2022dua}. All the other methods are evaluated with batch size of 8. 
 }
\end{table*}
\subsubsection{Evaluation with single distribution shift.}\label{sec:comparison_single_distribution_shift}
To evaluate the adaptation efficacy of our method and the baselines, we follow~\cite{schiappa2022large,yi2021benchmarking}, and apply each of the 12 corruption types to the entire validation set of all three datasets. We then adapt the networks to each of the 12 resulting corrupted validation sets.
We report adaptation results, averaged over the corruption types, in Table~\ref{tab:main-batchsize-1_8}. Comparing \textit{clean} and \textit{Source-Only}, we see the corruptions drastically deteriorate the performance of both CNN-based TANet and Video Swin Transformer.
On adapting the network, the baselines struggle in the online scenario and none of them attains large improvements over the un-adapted model for all three datasets.
By contrast, \OurMethod yields consistent and significant performance gains in the challenging and practical scenario where videos are received singly.

For a thorough comparison, we also evaluate the adaptation methods with a larger batch size of 8 in Table~\ref{tab:main-batchsize-1_8},   which is an easier setting for the baseline methods. The baseline methods in general have improvements in comparison to the case of batch size of 1. Note that results of \OurMethod with batch size of 1 in Table~\ref{tab:main-batchsize-1_8} already surpass the results of all baselines with batch size of 8 to a large margin. As \OurMethod accumulates the test statistics in an online manner instead of relying on statistics in a data batch, it does not require a large batch size for good adaptation performance. We further report the results of all the 12 corruption types for adaptation of TANet on UCF101 in Table~\ref{tab:resultsCorruptionWise_tanet_ucf}. The results indicate that for most of the corruption types, we outperform the baseline methods to a large margin. More results such as computational efficiency, adaptation on time-correlated data, and adaptation with train statistics from a different dataset can be found in the supplementary.

\begin{table}[!tb]
\small
\begin{tabular}{@{}M{0.8cm}M{0.8cm} M{0.8cm} M{0.8cm} M{0.7cm} M{0.7cm}M{0.8cm}@{}}
\toprule
Model & \multicolumn{3}{c}{TANet} & \multicolumn{3}{c}{Swin} \\
\cmidrule(r){2-4}\cmidrule{5-7}
Dataset & UCF101 & SSv2 & K400 &  UCF101 &  SSv2 & K400\\
\midrule
Source-Only & 55.41 & 26.93 & 39.98 & 79.62 & 44.31 & 49.48 \\
\midrule
NORM & 33.32 & 10.63 & 27.22  & - & - & - \\
DUA & 41.94 & 12.53 & 30.89 & - & - & -  \\
TENT & 31.32 & 10.68 & 27.25 & \underline{81.35} & \underline{44.58} & 49.46 \\
SHOT & 32.91 & 9.02  & 22.89 & 78.66 & 32.93 & 42.37 \\
T3A & \underline{53.32} & \underline{24.74} & \underline{38.86} & 81.02 & 43.54 & \underline{49.59} \\
\midrule
\OurMethod & \textbf{66.94} & \textbf{32.87} & \textbf{42.76} & \textbf{83.11} & \textbf{46.32} & \textbf{49.67} \\ 
\bottomrule
\end{tabular}
\vspace{-2mm}
\caption{
\label{tab:resultsMixedCorruptions}
Mean Top-1 Classification Accuracy (\%) with random distribution shifts. For each video in the test set, we randomly select 1 out of 13 distribution shifts (12 corruption types + \textit{original test set}). We run these experiments 3 times while shuffling the order of the videos and report the average results.}
\end{table}

\subsubsection{Evaluation with random distribution shift} \label{sec:comparison_random_distribution_shift}
We also evaluate the methods for online adaptation in a practical scenario where we assume that each video received has a random type of distribution shift. This scenario might happen when clients in different locations upload videos to the same platform. 
Specifically, for each of the videos in a sequence, we randomly select one of the 13 distribution shift cases (12 corruption types plus the case of \textit{no corruption}).

This setting is extremely challenging for our online method, since the corruption type changes in each iteration, and the distribution shift between training and test data becomes more complex.
This is reflected in the results in Table~\ref{tab:resultsMixedCorruptions}. For many combinations of the dataset and the backbone architecture, the baselines decrease the performance of the un-adapted model.
Our method consistently boosts the performance across datasets and architectures.
We attribute this robustness against changing corruption types to our technique of aggregating the statistics over multiple adaptation iterations.
This decreases the variation in the gradients computed from batches of data with different corruptions. 

\subsection{Ablation studies}
\label{sec:experiments_ablations}


\begin{table}[!tb]
\setlength\tabcolsep{10.0pt}

\small
\begin{tabular}{lcccc}
\toprule
Blocks  & 4 & 3, 4 & 2, 3, 4 & 1, 2, 3, 4 \\
\midrule
TANet & \underline{77.83} &  \textbf{78.20}  & 76.67 & 75.37 \\
Swin & 82.18 & \textbf{84.63} & \underline{84.47} & 84.40\\
\bottomrule
\end{tabular}
\vspace{-2mm}
\caption{
\label{tab:resultsLayers2Align}
Mean Top-1 Classification Accuracy (\%) over corruptions. We adapt TANet and Video swin tansformer on UCF101 by aligning feature maps from different blocks. \textit{block} refers to either convolutional \textit{bottleneck} in TANet or \textit{stage} in Swin transformer.
}
\end{table}

\subsubsection{The choice of feature maps to align}
\label{sec:choice_of_feature_map_to_align}
Aligning the distribution of the ultimate feature map of a deep network is a common practice in TTA, and in the broader field of domain adaptation~\cite{ganin2016domain,sun2017correlation,zellinger2017central,kojima2022cfa}. 
On the other hand, recent work hints that adapting multiple layers in the network~\cite{mirza2022dua,schneider2020norm,yin2020dreaming} might be a more powerful technique, at least when the adaptation is limited to statistics of the batch normalization layers.
To verify which of these approaches fares better in our task,
we run a series of experiments in which we adapt the TANet and Video Swin transformer to the corrupted UCF101 validation sets. We apply the alignment to the feature map outputs from different combinations of blocks. Here \textit{block} refers to either convolutional \textit{bottleneck} in TANet or \textit{stage} in Swin transformer. We repeat these experiments for four adaptation variants: either by the last block of the architecture, or by the last two, three, or all four blocks.
The results, presented in Table~\ref{tab:resultsLayers2Align}, suggest that an intermediate approach: aligning feature maps produced by the last two blocks, performs best.
We attribute this to the fact that leaving too many degrees of freedom during adaptation might lead to matching the distribution of the last layers, but without transferring feature semantics. 
On the other hand, some degree of freedom might be needed in the lower layers of the architecture for the network to learn to map appearance of the corrupted data to the feature space of the layers further in the computation graph, learned on the training data without corruption.

\begin{table}[!tb]
\small
\begin{tabular}{lcccc}
\toprule
 & Statistics & UCF101 & SSv2 & K400 \\
\midrule
Source-Only & \_ &51.35 & 24.31 & 37.16 \\
\OurMethod &BNS& 73.20 & 35.51 & 45.30  \\
\OurMethod  &Src-Computed& \textbf{78.20} & \textbf{37.97} & \textbf{48.69} \\ 
\bottomrule
\end{tabular}
\vspace{-2mm}
\caption{
\label{tab:resultsRunningStats_new}
Mean Top-1 Classification Accuracy (\%) over all corruptions on the UCF101 dataset, while using different kinds of source statistics for alignment. We use the TANet backbone for these experiments. $BNS$ refers to the statistics stored in the batch normalization layers. $Src-Computed$ refers to the statistics which we calculate from the source data for distribution alignment.  
}
\end{table}
\subsubsection{Statistics stored in normalization layers}
\label{sec:statistics_norm_layers}
For feature distribution alignment, our method requires feature means and variances computed on the training data.
When training data is no longer available, these statistics could be computed on other data with the similar distribution.
For architectures that contain batch normalization layers, using the running means and variances, accumulated at these layers during training, represents a convenient alternative.
However, these statistics might misrepresent true means and variances of the data, as reported by Wu et al.~\cite{wu2021rethinking}.
To investigate how their inaccuracy affects performance, we compare the results attained by using them as adaptation targets to those obtained by using statistics computed on the training set. 
We present the results in Tab~\ref{tab:resultsRunningStats_new}. They confirm that relying on the running mean and variance yields performance slightly lower than computing the statistics from scratch, but still higher than that of the baselines. This demonstrates that our \OurMethod can also only rely on batch norm statistics in architectures with BN layers, at a modest performance penalty.

\begin{table}[!tb]
\setlength\tabcolsep{4.0pt}
\small

\begin{tabular}{lcccccc}

\toprule
Views & 1 & 2 & 2 & 3 & 4 & 5 \\
Pred. Cons. &\xmark&\xmark&\cmark&\cmark&\cmark&\cmark\\

\midrule
 Acc & 75.57 &  77.46  & 78.20 & 78.24 & 78.25 & 78.25  \\

\bottomrule
\end{tabular}
\vspace{-2mm}
\caption{
\label{tab:resultsNoViews}
Mean Top-1 Classification Accuracy (\%) over all corruption types for UCF101 by using the TANet backbone. We perform ablation study on different number of temporal views and the prediction consistency regularization.
}
\end{table}
\subsubsection{Number of views and prediction consistency}
\label{sec:num_views_pred_consis}
In temporal augmentation, we compute test statistics on temporally augmented views of the input and enforce prediction consistency among the views. We verify how much performance gain these design choices yield in Table~\ref{tab:resultsNoViews}. Without prediction consistency, sampling two temporally augmented views (77.46\%) brings around two percent improvement in comparison to only one view (75.57\%). Applying the prediction consistency regularization among the two views further adds 0.74\% improvement. Sampling more than two augmented views adds little benefit. We set the number of views to two for better adaptation efficiency. 


\begin{table}[!tb]
\setlength\tabcolsep{4.0pt}
\small
\begin{tabular}{lccccc}
\toprule
sampling  & uniform & dense & uniform & dense & total \\

strategy&random&random&equidistant&equidistant&random\\
\midrule
Accuracy &  77.68 &  76.93  & \textbf{78.20} & 77.10 & 77.50 \\
\bottomrule
\end{tabular}
\vspace{-2mm}
\caption{
\label{tab:resultSampling}
Mean Top-1 Classification Accuracy (\%) over all corruptions on UCF101 using TANet. We perform ablation study on the frame sampling strategies for temporally augmented views. }

\end{table}



\subsubsection{Temporal sampling strategies}
\label{sec:temp_samp_strategies}
Our method relies on temporal augmentation to generate multiple views of the input video sequence and enforce consistency of their predictions.
Common temporal sampling methods can be found in video recognition literature~\cite{tran2015learning,carreira2017quo,feichtenhofer2020x3d,wang2016temporal}. We embed the most representative of these techniques in our adaptation algorithm and evaluate the resulting performance. \textit{Uniform} and \textit{dense} are common strategies for video segment selection. \textit{Random} and \textit{equidistant} are strategies of sampling frames in each video segment. \textit{total random} refers to completely randomly sampling frames from the entire video.

The results in Table~\ref{tab:resultSampling} show that our \OurMethod generalizes well to different types of frame sampling strategies, as they all demonstrate clear performance boost in comparison to the case of one view (75.57\%). The \emph{uniform-equidistant} approach perform best. We hypothesize that this stems from the fact that this sampling technique keeps the interval between the video frames constant, while yielding frame sequences that span the entire video sequence, which leads to more accurate statistics of the overall video content.

\subsubsection{Continuous adaptation}
\label{sec:continuous_adapt}
We check the capacity of the methods to re-adapt to the un-corrupted data. 
We perform an experiment in the continuous adaptation scenario.
When sequentially feeding the test data, we periodically switch the corruption `on' and `off' around every five hundred test videos. We use the Gaussian noise as the corruption technique.

\begin{figure}[t]
\centering
\includegraphics[width=\linewidth]{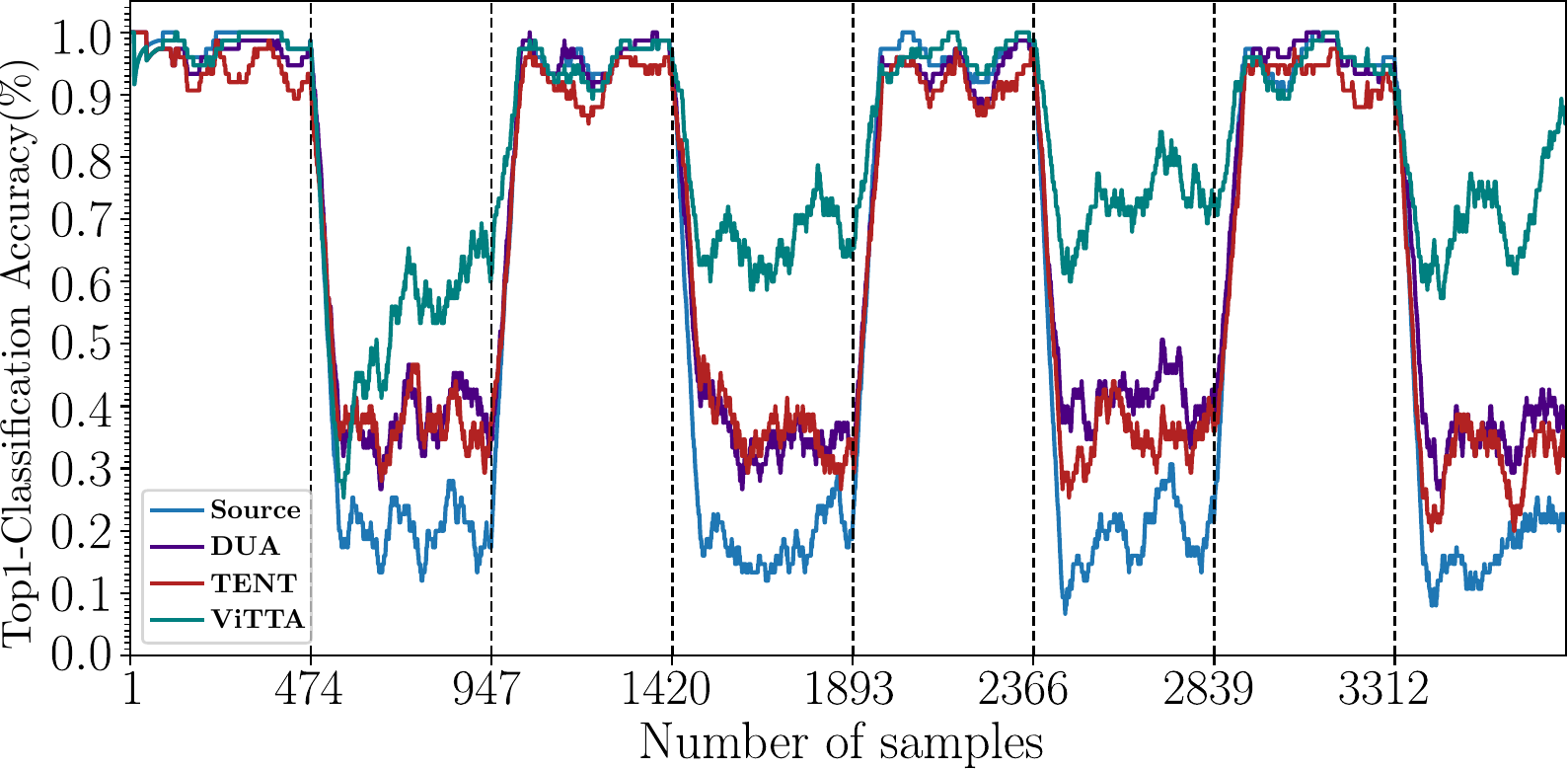}
\caption{
\label{fig:resultsOnOff}
Top-1 Classification Accuracy~(\%) for a sliding window of $75$ samples in a continuous adaptation scenario, where we switch alternatively between Gaussian Noise corruption and the Clean test set of UCF101, we compare with the two best performing baselines \textit{DUA} and \textit{TENT} to avoid clutter. \textit{Source} refers to performance of the pre-trained model without adaptation.}
\end{figure}

As seen in Fig.~\ref{fig:resultsOnOff}, \OurMethod performs constantly the best on corrupted periods. Both our approach and \textit{DUA} can recover the original performance on the un-corrupted periods, while \textit{TENT} performs slightly worse when adapting to the un-corrupted data. \textit{DUA} only corrects the batch norm statistics without updating model parameters. It keeps the knowledge on the un-corrupted but has limited gain in corrupted periods. \textit{TENT} updates the model with an entropy minimization loss and has even slightly worse performance than \textit{Source-Only} in un-corrupted periods. In comparison, our method updates the entire model and still quickly restores the performance in un-corrupted periods, demonstrating the fast reactivity. As our method accumulates the target statistics in an online manner, it also has gradually improved performance when going through the four corrupted periods. 

\section{Conclusion}
\label{sec:conclusion}
We address the problem of test-time adaptation of video action recognition models against common corruptions, We propose a video-tailored method that aligns the training statistics with the online estimates of target statistics. To further boost the performance, we enforce prediction consistency among temporally augmented views of a video sample. We benchmark existing TTA techniques on three action recognition datasets, with 12 common image- and video-specific corruptions. Our proposed method \OurMethod performs favorably in evaluation of both, single corruption and the challenging random corruption scenario. Furthermore, it demonstrates fast reactivity on adaptation performance, faced with periodic change of distribution shift. 

\myparagraph{Acknowledgements} 
We gratefully acknowledge the financial support by the Austrian Federal Ministry for Digital and Economic Affairs, the National Foundation for Research, Technology and Development and the Christian Doppler Research Association. This work was also partially funded by the FWF Austrain Science Fund Lise Meitner grant (M3374).


\appendix
\section*{Supplementary}
For further insights into our \OurMethod, we provide implementation details of TANet and Video Swin Transformer (Sec.~\ref{sec:specifics_vid_models}), introduction of the corruptions used in evaluation (Sec.\ref{sec:corr_intro}) and additional experimental results (Sec.~\ref{sec:additional_results}). 

In Sec.~\ref{sec:computational_efficiency}, we compare the computation efficiency of \OurMethod with baseline methods. In Sec.~\ref{sec:eval_time_correlated_samples}. we report adaptation performance on time-correlated data. 
In Sec.~\ref{sec:adapt_part_of_test}, we report the performance of adaptation to only a part of the test set to validation the adaptation efficiency of \OurMethod. 
In Sec.~\ref{sec:train_stats_from_different_dataset}, we compare the performance of adaptation to a test set using train statistics from different datasets. 
In Sec.~\ref{sec:momentum_moving_average}, we perform an ablation study on the choice of momentum for moving average. Furthermore, we report standard deviation of evaluations with random distribution shift in Sec.~\ref{sec:random_distribution_shift}, and report the results of all corruptions for the single distribution shift evaluation in Sec.~\ref{sec:results_of_all_corr}.

\section{Implementation Details} \label{sec:specifics_vid_models}
\subsection{TANet}
\myparagraph{Training}. The pre-trained \texttt{TAM-R50} models of TANet~\cite{liu2021tam} based on ResNet50~\cite{he2016deep} for K400 and SSv2 are from the model zoo of TANet\footnote{\url{https://github.com/liu-zhy/temporal-adaptive-module}}. We use the models that are trained with 8 frames sampled from each video for both datasets. 

We train \texttt{TAM-R50} for UCF101 with weights initialized from the model pre-trained on K400. Following~\cite{liu2021tam}, we resize the shorter size of the frame to 256, and apply the multi-scale cropping and random horizontal flipping as data augmentation. The cropped frames are resized to 224$\times$224 for training. 
We use a batch size of 24 and train for 50 epochs using SGD with momentum of 0.9, weight decay of $1e-4$, and initial learning rate of $1e-3$. We also the uniform sampling strategy and sample 16 frames per video. 

\myparagraph{Inference}. During inference, we resize the shorter side of the frame to 256. For efficient implementation, we take the center crop of size 224$\times$224 in the spatial dimensions. For frame sampling, we apply uniform sampling strategy to sample one clip from each test video. The clip length is 8 for K400 and SSv2, and 16 for UCF101. 

\subsection{Video Swin Transformer}
\myparagraph{Training.} We use the pre-trained models of the Video Swin Transformer~\cite{liu2022video} from the model zoo of Video Swin Transformer\footnote{\url{https://github.com/SwinTransformer/Video-Swin-Transformer}}. The backbone is \texttt{Swin-B} from Swin Transformer~\cite{liu2021swin}. For K400, we use the \texttt{Swin-B} model that is trained with weight initialization from pre-training on ImageNet-1K~\cite{deng2009imagenet}. For SSv2, the model is trained with weight initialization from pre-training on K400.  


For UCF101, we train \texttt{Swin-B} model with the weight initialization from K400. Following~\cite{liu2022video}, the model is trained using an AdamW~\cite{kingma2014adam} optimizer with a cosine decay learning rate scheduler. The initial learning rate is set to $3e-5$. We train with batch size of 8 for 30 epochs. A clip of 32 frames (size 224$\times$224) with a temporal stride of 2 is sampled from each video. 

\myparagraph{Inference}. During inference, we take the center crop of size 224$\times$224. For frame sampling, we apply uniform sampling strategy to sample one clip of 16 frames from each test video. 

\section{Corruption Types}\label{sec:corr_intro}
We evaluate on 12 types of corruptions in video acquisition and video processing, proposed in~\cite{schiappa2022large,yi2021benchmarking} that benchmark the robustness of spatio-temporal models. We prepare the corrupted videos for UCF101~\cite{soomro2012ucf101}, Something-something v2 (SSv2)~\cite{goyal2017something} and Kinetics 400 (K400)~\cite{kay2017kinetics} with implementations from these two works. We give short descriptions of the 12 corruptions in the following. 

\textit{Gaussian Noise} could appear due to low-lighting conditions or sensor limitation during video acquisition. \textit{Pepper Noise} and \textit{Salt noise} simulates disturbance in image signal with densely occurring black pixels and white pixels. \textit{Shot noise} captures the electronic noise caused by the discrete nature of light. In the implementation, it is approximated with a Poisson distribution. \textit{Zoom Blur} occurs when the camera moves towards an object rapidly. \textit{Impulse noise} simulates corruptions caused by the defect of camera sensor. \textit{Defocus Blur} happens when the camera is out of focus. \textit{Motion Blur} is caused by destabilizing motion of camera. \textit{Jpeg} is a lossy image compression format which introduces compression artifacts. \textit{Contrast} corruption is common in video acquisition due to changing contradiction in luminance and color of the scene that is captured. \textit{Rain} corruption simulates a rainy weather condition during video acquisition. \textit{H265 ABR Compression} corruption simulates compression artifacts when video compression is performed using the popular H.265 codec with an average bit rate. 

\begin{figure*}
\includegraphics[width=0.8\columnwidth]{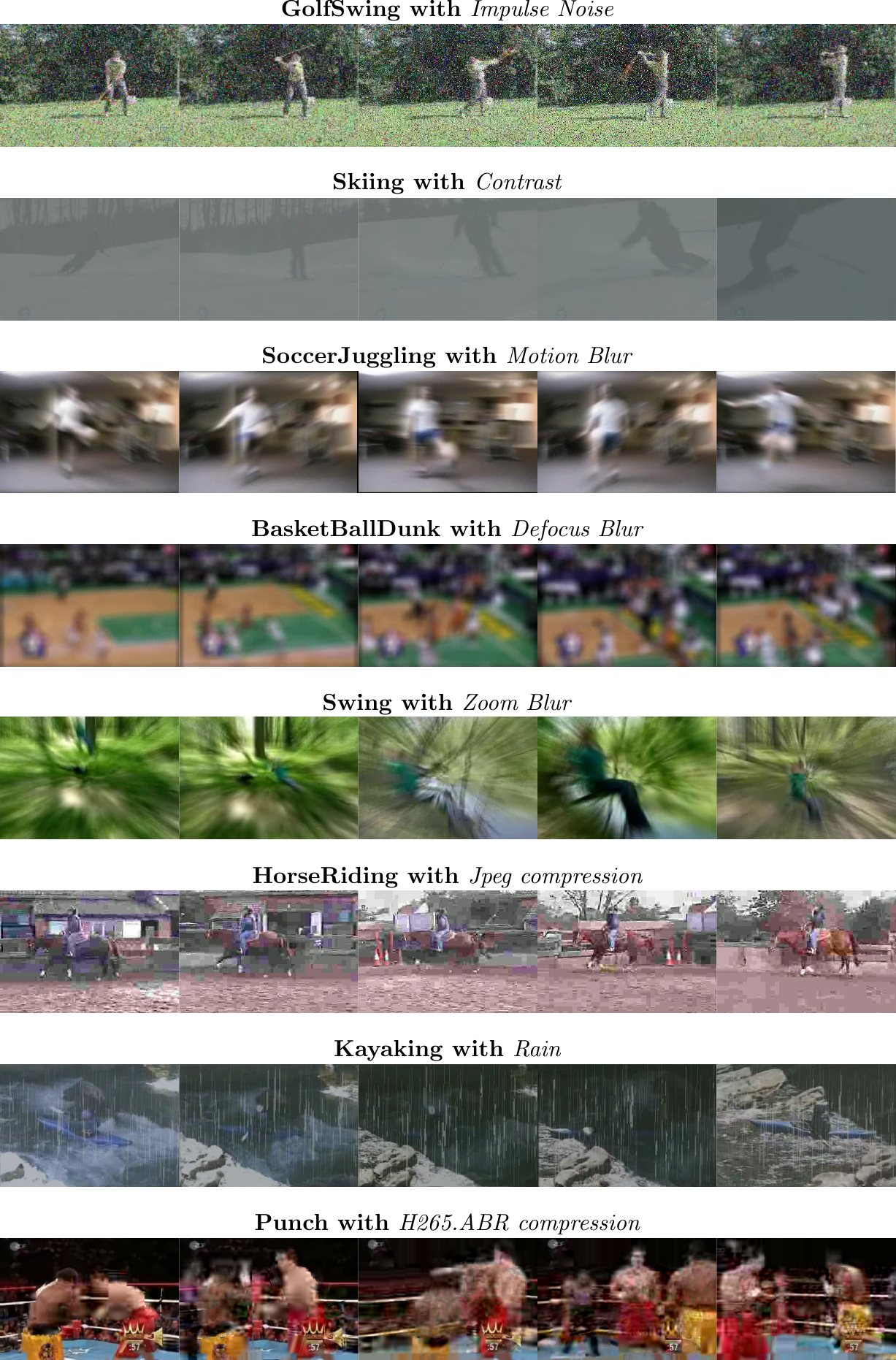}
\caption{
\label{fig:corr_visual}
Samples of action videos with corruptions.  Best viewed on screen.
}
 
\end{figure*}

We visualize some samples of action videos with corruptions from UCF101 in Fig.~\ref{fig:corr_visual}. 

\section{Additional Results}\label{sec:additional_results}
\subsection{Computational Efficiency}\label{sec:computational_efficiency}
We report Memory (MB), latency (s) and performance of TANet on UCF101 (Nvidia A6000 + AMD EPYC 7413) in Tab.~\ref{tab:memory_latency}. ViTTA offers an excellent performance/computation tradeoff.

\begin{table}[!tb]
\small
\begin{tabular}{ccccccc} 
\toprule
method & batch & views & block & mem. & latency & acc\\
\midrule
ViTTA & 1 & 2 & 3,4 & 6500 & 0.159 & \textbf{78.20}\\
ViTTA & 1 & 2 & 4 & 6236 & 0.146 & 77.83\\
ViTTA & 1 & 1 & 3,4 & 4601 & 0.110 & 75.57\\
ViTTA & 1 & 1 & 4 & 4469 & 0.098 & 75.43\\
\midrule
TENT & 8 & - & - & 16482 & 0.456 & \textbf{72.92}\\
NORM & 8 & - & - & 3410 & 0.262 & 65.77\\
SHOT & 8 & - & - & 16483 & 0.533 & 65.54\\
DUA & 1 & - & - & 4785 & 0.830 & 55.34\\
T3A & 8 & - & - & 3410 & 0.151 & 54.17\\
T3A & 1 & - & - & 2702 & 0.042 & 54.17\\
NORM & 1 & - & - & 2702 & 0.056 & 51.59\\
TENT & 1 & - & - & 4339	& 0.081 & 51.58\\
SHOT & 1 & - & - & 4339	& 0.089 & 51.20\\
\bottomrule
\end{tabular}
\caption{
\label{tab:memory_latency}
Computational efficiency.}
\end{table}

\subsection{Evaluation on Time-Correlated Data}\label{sec:eval_time_correlated_samples}
In Tab.~\ref{tab:eval_time_correlated_samples}, we run experiments in adaptation to sequences of videos in the order given in the original validation lists, where videos of the same class are listed together. This makes them highly correlated. On UCF, even clips of the same scene/video are listed together. ViTTA still outperforms all baselines by a significant margin. 

\begin{table}[!tb]
\small
    \begin{tabular}{M{0.8cm}M{0.8cm}M{0.6cm}M{0.6cm}M{0.6cm}M{0.6cm}M{0.6cm}M{0.6cm}}
    \toprule
    model & \multirow{2}*{batch} & \multicolumn{3}{c}{TANet}  & \multicolumn{3}{c}{Swin} \\
    \cmidrule(r){3-5}\cmidrule{6-8}
    dataset & ~ & UCF & SSv2 & K400 &  UCF &  SSv2 & K400\\
    \midrule
    source & - & 51.35 & 24.31 & 37.16 & 78.48 & 42.18 & 47.17 \\
    \midrule
    NORM & 1 & 17.14 & 3.77 & 8.72 & - & - & - \\
    DUA & 1 & 24.83 & 6.45 & 13.23 & - & - & - \\
    TENT & 1 & 17.14 & 3.79 & 8.77 & 79.35 & 33.22 & 45.64 \\
    SHOT & 1 & 17.30 & 3.71 & 5.79 & 68.96 & 20.58 & 31.04 \\
    T3A & 1 & 53.19 & 24.14 & 37.72 & 80.18 & 41.64 & 48.20 \\
    \textbf{ViTTA} & 1 & \textbf{71.40} & \textbf{31.45} & \textbf{45.04} & \textbf{83.48} & \textbf{42.98} & \textbf{48.97} \\
    \midrule
    NORM & 8 & 52.87 & 9.83 & 27.52 & - & - & - \\
    TENT & 8& 53.91 & 10.24 & 27.43 &  81.11 & 42.81 & 47.85 \\
    SHOT & 8& 52.90 & 9.85 & 26.27 & 77.83 & 39.52 & 46.05 \\
    T3A & 8& 53.47 & 24.17 & 37.75 & 80.31 & 41.69 & 48.22 \\
    \textbf{ViTTA} & 8 & \textbf{74.19} & \textbf{33.31} & \textbf{46.70} & \textbf{83.92} & \textbf{48.25} & \textbf{53.97} \\
    \bottomrule
    \end{tabular}
\caption{
\label{tab:eval_time_correlated_samples}
Mean Top-1 Classification Accuracy (\%) over all corruption types on UCF101, SSv2 and K400 datasets. Adaptation on time correlated videos.}
\end{table}

\subsection{Adapting To Only A Part of Test Videos}\label{sec:adapt_part_of_test}
During online adaptation, we estimate the test set statistics by continuous exponential moving average of statistics computed on a stream of test videos, and use these estimated statistics for alignment. Here we study the performance of our \OurMethod when adapting to only the first portion of all test videos in the sequence. 

We denote the number of videos in the test set as $N_T$ and the percentage of videos that we adapt to as $p\cdot 100\%$. The sequence of test videos is $\inpt_1, ..., \inpt_i, ..., \inpt_{N_T}$, where $i$ is the video index. 
We perform online adaptation only on the first $p\cdot 100\%$ of test videos in the sequence. Consequently, $N_{adapt}=p\cdot N_T$ denotes the absolute number of videos used for adaptation. 

\begin{figure}[t]
\centering
\includegraphics[width=\linewidth]{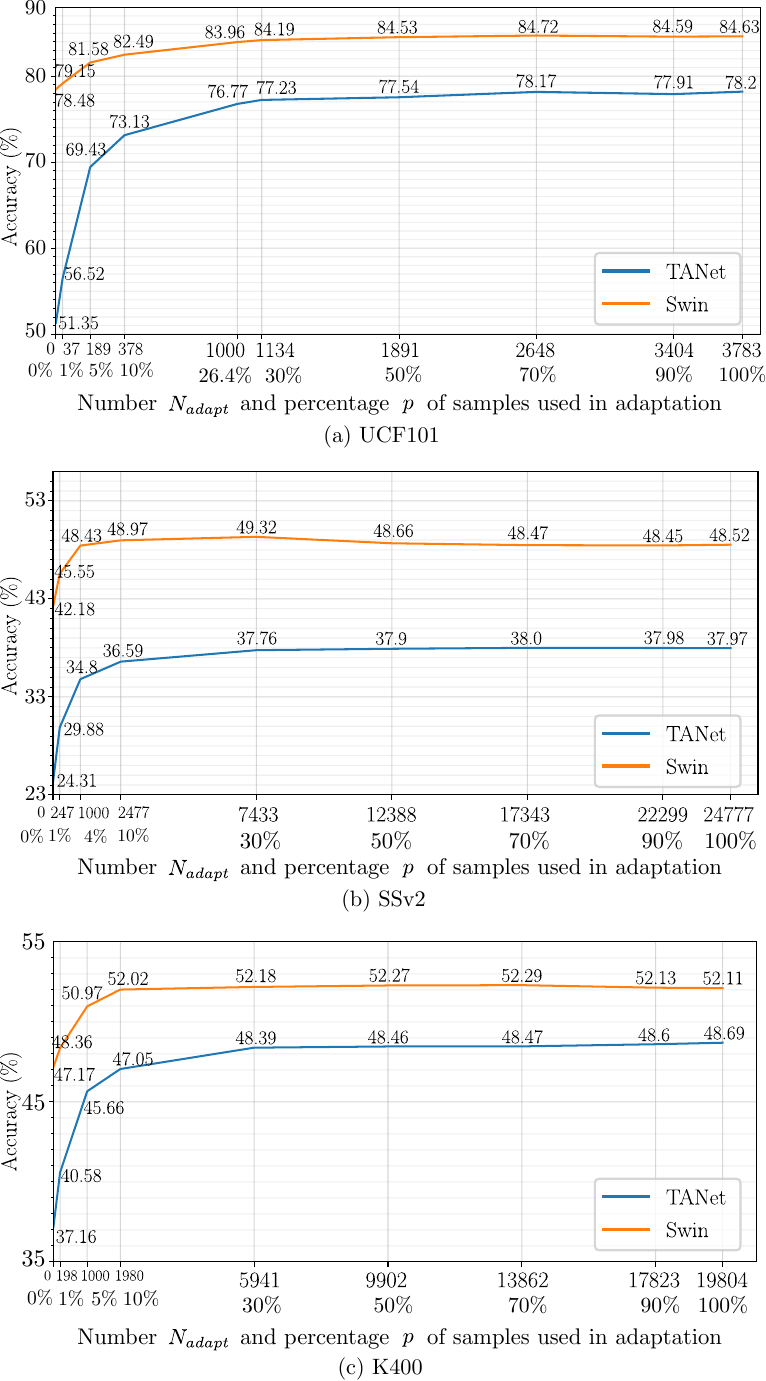}
\caption{
\label{fig:adapt_ratio}
Mean Top-1 Classification Accuracy (\%) over all corruptions of adaptation to only the first $N_{adapt}$ test videos ($p\%$) in the sequence. For videos with index $i\leq N_{adapt}$, we test right after adapting to it. For videos with index $i> N_{adapt}$, we test directly without any further adaptation. We vary $p$ from 0\% (\textit{Source-Only, $N_{adapt}=0$}) to 100\% (full adaptation, $N_{adapt}=N_T$), and report the performance of TANet and Video Swin Transformer on three datasets in (a) UCF101 (b) SSv2 and (c) K400. }\end{figure}

Specifically, for videos with index $i\leq N_{adapt}$, we test right after adapting to them. For videos with index $i> N_{adapt}$, we test directly without any further adaptation. We vary $p$ from 0\% (\textit{Source-Only, $N_{adapt}=0$}) to 100\% (full adaptation, $N_{adapt}=N_T$), and report the performance of TANet and Video Swin Transformer on three datasets in Fig.~\ref{fig:adapt_ratio}. 

We notice that on all datasets, when $N_{adapt}$ is below 1000, the adaptation performance improves fast with $N_{adapt}$ increasing. After $N_{adapt}$ reaches 1000, increasing $N_{adapt}$ leads to stable and saturated adaptation performance.
Adapting only to the first 1000 videos leads to performance drop of less than 3\% in comparison to the full adaptation case. Note that $N_{adapt}=1000$ corresponds to $p$ of 26.4\% on the small UCF101 and only 4\% $\sim$ 5\% on the large-scale SSv2 and K400. This indicates the adaptation efficiency of our \OurMethod. 

\subsection{Train Statistics From A Different Dataset}\label{sec:train_stats_from_different_dataset}
For feature distribution alignment, our \OurMethod requires feature means and variances computed on the training data. When training data is no longer available, these statistics could be computed on other data with similar distribution. We compare the performance of adaptation to a test set using train statistics from different datasets in Table~\ref{tab:ablation_train_stats}. Intuitively, using the training statistics from the same dataset leads to the best adaptation performance. Adaptation with train statistics on clean data leads to performance boost upon \textit{Source-Only}, even if the train statistics are from a different dataset. 
As UCF101 and K400 are both generic action datasets in user video style and have similar statistics, using the train statistics from one to adapt to the test set of the other leads to slight performance drop. SSv2 is a egocentric motion-based dataset and has significantly different distribution than UCF101 and K400. Adaptation across SSv2 and UCF101 or K400 leads to less satisfying performance improvement.

\begin{table}[!tb]
\setlength\tabcolsep{10.0pt}

\small
\begin{tabular}{M{1cm}M{1.3cm}M{0.6cm}M{0.5cm}M{0.5cm}}
\toprule
\multirow{2}*{Test Set}  & \multirow{2}*{Source-Only} & \multicolumn{3}{c}{Train Statistics} \\
\cmidrule(l){3-5}
~ & ~ & UCF101 & SSv2 & K400 \\
\midrule
UCF101 & 51.35 &  \textbf{78.20}  & 65.64 &  \underline{74.53} \\
SSv2 & 24.31 & \underline{30.34} & \textbf{37.97} & 29.41 \\
K400 & 37.16 & \underline{46.94} & 38.46 & \textbf{48.69}  \\
\bottomrule
\end{tabular}
\caption{
\label{tab:ablation_train_stats}
Mean Top-1 Classification Accuracy (\%) over all corruptions of adaptation to a test set with train statistics from different datasets. We perform adaptation on TANet in combination of different train and test sets across UCF101, SSv2 and K400.
}
\end{table}

\subsection{Momentum of Moving Average}\label{sec:momentum_moving_average}
We approximate the test set statistics by exponential moving averages of statistics computed on consecutive test videos.
Recall Eqs.~(4) and~(5) of the main manuscript:
\begin{equation} \label{eq:runnin_mean}
    {\mean_{\layer}}^{(i)}(\param)=\alpha\cdot \mean_{\layer}(\inpt_i;\param) + (1-\alpha)\cdot {\mean_{\layer}}^{(i-1)}(\param),
\end{equation}
\begin{equation} \label{eq:runnin_var}
    {\var_{\layer}}^{(i)}(\param)=\alpha\cdot \var_{\layer}(\inpt_i;\param) + (1-\alpha)\cdot {\var_{\layer}}^{(i-1)}(\param).
\end{equation}
Here we have the hyperparameter $\alpha$ and, consequently, $1-\alpha$ is the momentum (erratum for line 408 in the main paper: $\alpha$ is a hyperparameter, not the momentum). As we evaluate with the small batch size of 1, a large momentum will result in a steady update of statistics. We set $\alpha$ to 0.1, which corresponds to a common choice of large momentum of 0.9. We study other choices of the momentum value and report the performance in Table~\ref{tab:momentum_mvg}.

\begin{table}[!tb]
\setlength\tabcolsep{10.0pt}

\small
\begin{tabular}{M{1.2cm}M{0.5cm}M{0.5cm}M{0.5cm}M{0.5cm}M{0.5cm}}
\toprule
Momentum $1-\alpha$  & 0.99 & 0.95 & 0.9 & 0.85 & 0.8 \\
\midrule
UCF101 & 56.94 &  64.70  & \textbf{78.20} &  \underline{77.53} &  76.80\\
SSv2 & 34.81 & \underline{37.53} & \textbf{37.97} & \underline{37.53} & 36.13 \\
K400 & 45.68 & \underline{48.15} & \textbf{48.69} & 47.47 & 46.13 \\
\bottomrule
\end{tabular}
\caption{
\label{tab:momentum_mvg}
Mean Top-1 Classification Accuracy (\%) over all corruptions with different momentum values ($1-\alpha$) for moving average. We report adaptation performance of TANet on three action datasets. 
}
\end{table}

When the momentum is too large, there is a slow convergence of statistics. The moving average is dominated by statistics of features in the past, which are outdated as the model evolves~\cite{wu2021rethinking}. As shown in Table~\ref{tab:momentum_mvg}, this slow convergence has a big impact on a small video dataset like UCF101 (3783 validation videos), resulting in a drastic performance drop. In the meanwhile, on large video datasets like SSv2 (24K validation videos) and K400 (20K validation videos), the performance drop due to slow convergence is less severe. 
When the momentum is too small, the estimated statistics are dominated by the recent samples and do not represent the whole population, leading to performance degradation.

\subsection{Evaluation with Random Distribution Shift}\label{sec:random_distribution_shift}
In Sec.4.3.2 in the main paper, we evaluated in the scenario where we assume that each video received has a random type of distribution shift. For each of the videos in a sequence, we randomly selected one of the 13 distribution shifts (12 corruption types, plus the case of \textit{no corruption}). We ran the experiments 3 times while shuffling the order of the videos. 
In addition to the average results (reported in Table~3 in the main paper), here we also report the standard deviation in Table~\ref{tab:resultsMixedCorruptions}. Shuffling the video and corruption order leads to only minor variation in the performance. Considering the standard deviation, our \OurMethod still constantly outperforms baseline methods in all settings. 

\begin{table}[!tb]
\small
\begin{tabular}{@{}M{0.8cm}M{0.8cm} M{0.8cm} M{0.8cm} M{0.7cm} M{0.7cm}M{0.8cm}@{}}
\toprule
Model & \multicolumn{3}{c}{TANet} & \multicolumn{3}{c}{Swin} \\
\cmidrule(r){2-4}\cmidrule{5-7}
Dataset & UCF101 & SSv2 & K400 &  UCF101 &  SSv2 & K400\\
\midrule
source-only & 55.41 & 26.93 & 39.98 & 79.62 & 44.31 & 49.48 \\
\midrule
NORM & 33.32$\pm$ 0.09 & 10.63$\pm$ 0.09 & 27.22$\pm$ 0.12  & - & - & - \\
DUA & 41.94$\pm$ 0.40 & 12.53$\pm$ 0.19 & 30.89$\pm$ 0.18 & - & - & -  \\
TENT & 31.32$\pm$ 0.06 & 10.68$\pm$ 0.11 & 27.25$\pm$ 0.09 & \underline{81.35}$\pm$ 0.25 & \underline{44.58}$\pm$ 1.96 & 49.46$\pm$ 0.08 \\
SHOT & 32.91$\pm$ 0.16 & 9.02$\pm$ 0.12  & 22.89$\pm$ 0.12 & 78.66$\pm$ 0.21 & 32.93$\pm$ 0.40 & 42.37$\pm$ 1.77 \\
T3A & \underline{53.32}$\pm$ 0.21 & \underline{24.74}$\pm$ 0.12 & \underline{38.86}$\pm$ 0.14 & 81.02$\pm$ 0.14 & 43.54$\pm$ 0.09 & \underline{49.59}$\pm$ 0.08 \\
\midrule
\OurMethod & \textbf{66.94}$\pm$ 0.31 & \textbf{32.87}$\pm$ 0.08 & \textbf{42.76}$\pm$ 0.07 & \textbf{83.11}$\pm$ 0.35 & \textbf{46.32}$\pm$ 0.11 & \textbf{49.67}$\pm$ 0.01 \\ 
\bottomrule
\end{tabular}
\caption{
\label{tab:resultsMixedCorruptions}
Top-1 Classification Accuracy (\%) with random distribution shifts. For each video in the test set, we randomly select 1 out of 13 distribution shifts (12 corruption types + \textit{original test set}). We run these experiments 3 times while shuffling the order of the videos and report the mean and standard deviation.}
\end{table}





\subsection{Evaluation with Single Distribution Shift: Results of All Corruptions}\label{sec:results_of_all_corr}
In Table~1 of the main paper, for evaluation with single distribution shift, we report the average performance on all corruption types. 
Here we report the detailed results of all the 12 corruption types for TANet (Table~\ref{tab:corr_results_tanet}) and Video Swin Transformer (Table~\ref{tab:corr_results_swin}) on the three datasets. 

\begin{table*}[!tb]
\centering
\small
\begin{tabular}{lM{0.75cm}M{0.75cm}M{0.75cm}M{0.75cm}M{0.75cm}M{0.75cm}M{0.75cm}M{0.75cm}M{0.75cm}M{0.75cm}M{0.75cm}M{0.75cm}M{0.75cm}}
\toprule
\multicolumn{14}{c}{TANet UCF101} \\
\midrule
corruptions & gauss & pepper & salt & shot & zoom  & impulse &  defocus & motion &  jpeg & contrast & rain & h265.abr & avg\\
\midrule
Source-Only & 17.92 & 23.66 & 7.85 &  72.48 & 76.04 & 17.16 & 37.51 & 54.51 & 83.40 & 62.68 & 81.44 & 81.58 & 51.35 \\
\midrule
NORM & 45.23 & 42.43 & 27.91 & 86.25  & \underline{84.43} & 46.31 & 54.32 & 64.19 & 89.19 & 75.26 & 90.43 & 83.27 & 65.77\\
DUA & 36.61 & 33.97 & 22.39 & 80.25 & 77.13 & 36.72 & 44.89 & 55.67 & 85.12 & 30.58 & 82.66 & 78.14 & 55.34 \\
TENT & \underline{58.34} & \underline{53.34} & \underline{35.77} & \underline{89.61} & \textbf{87.68} & \underline{59.08} & \underline{64.92} & \underline{75.59} & \underline{90.99} & \underline{82.53} & \underline{92.12} & \textbf{85.09} & \underline{72.92} \\
SHOT & 46.10 & 43.33 & 29.50 & 85.51  & 82.95 & 47.53 & 53.77 & 63.37 & 88.69 & 73.30 & 89.82 & 82.66 &  65.54\\
T3A & 19.35 & 26.57 & 8.83 & 77.19  & 79.38 & 18.64 & 40.68 & 58.61 & 86.12 & 67.22 & 84.0 & 83.45 & 54.17\\
\midrule
\OurMethod & \textbf{71.37} & \textbf{64.55} & \textbf{45.84} & \textbf{91.44}  & \textbf{87.68} & \textbf{71.90} & \textbf{70.76} & \textbf{80.32} & \textbf{91.70} & \textbf{86.78} & \textbf{93.07} & \underline{84.56} & \textbf{78.33}\\
\bottomrule
\end{tabular}

\small  
\begin{tabular}{lM{0.75cm}M{0.75cm}M{0.75cm}M{0.75cm}M{0.75cm}M{0.75cm}M{0.75cm}M{0.75cm}M{0.75cm}M{0.75cm}M{0.75cm}M{0.75cm}M{0.75cm}}
\toprule
\multicolumn{14}{c}{TANet SSv2} \\
\midrule
corruptions & gauss & pepper & salt & shot & zoom  & impulse &  defocus & motion &  jpeg & contrast & rain & h265.abr & avg\\
\midrule
Source-Only & 14.29 & 16.36 & 7.83 &  34.73 & 45.72 & 14.39 & 27.83 & 26.92 & 25.21 & 23.24 & 41.10 & 14.07 & 24.31 \\
\midrule
NORM & 15.99 & 16.95 & 12.87 & 40.14  & 45.30 & 17.06 & 34.53 & 32.47 & 41.77 & 23.05 & 42.18 & 15.89 & 28.18\\
DUA & 7.45 & 8.95 & 6.45 & 24.0 & 29.92 & 8.0 & 20.71 & 19.48 & 27.62 & 6.38 & 25.0 & 10.48 & 16.20 \\
TENT & \underline{20.52} & \underline{20.74} & \underline{15.45} & \underline{44.01} & \underline{47.11} & \underline{21.34} & \underline{38.10} & \underline{35.87} & \underline{45.26} & \underline{27.42} & \underline{45.70} & \underline{17.36} & \underline{31.57} \\
SHOT & 17.67 & 18.16 & 13.85 & 37.49  & 43.25 & 17.85 & 32.59 & 31.71 & 39.37 & 22.28 & 39.0 & 16.41 &  27.47\\
T3A & 15.23 & 16.49 & 7.69 & 33.89  & 44.26 & 15.48 & 27.69 & 27.17 & 29.08 & 23.34 & 39.71 & 13.37 & 24.45\\
\midrule
\OurMethod & \textbf{34.69} & \textbf{26.69} & \textbf{16.15} & \textbf{48.58}  & \textbf{49.26} & \textbf{35.54} & \textbf{44.05} & \textbf{40.93} & \textbf{48.37} & \textbf{42.05} & \textbf{50.95} & \textbf{19.57} & \textbf{38.07}\\
\bottomrule
\end{tabular}

\small  
\begin{tabular}{lM{0.75cm}M{0.75cm}M{0.75cm}M{0.75cm}M{0.75cm}M{0.75cm}M{0.75cm}M{0.75cm}M{0.75cm}M{0.75cm}M{0.75cm}M{0.75cm}M{0.75cm}}
\toprule
\multicolumn{14}{c}{TANet K400} \\
\midrule
corruptions & gauss & pepper & salt & shot & zoom  & impulse &  defocus & motion &  jpeg & contrast & rain & h265.abr & avg\\
\midrule
Source-Only & 28.52 & 26.14 & 15.97 &  57.89 & 39.66 & 29.51 & 47.01 & 52.79 & 60.60 & 25.69 & 48.23 & 13.91 & 37.16 \\
\midrule
NORM & 32.55 & 30.09 & 21.67 & 58.22  & 43.44 & 33.25 & 43.99 & 50.74 & 60.28 & 28.19 & 57.14 & 12.29 & 39.32\\
DUA & 24.26 & 22.25 & 14.68 & 50.74 & 35.10 & 25.04 & 38.15 & 43.41 & 53.29 & 18.32 & 48.04 & 9.31 & 31.88 \\
TENT & \underline{34.16} & \underline{31.54} & \underline{22.64} & \underline{60.29} & \underline{46.03} & \underline{34.88} & 45.54 & 52.50 & \underline{61.87} & \underline{32.50} & \underline{59.21} & 12.44 & \underline{41.13} \\
SHOT & 32.24 & 29.91 & 21.76 & 55.72  & 40.26 & 32.95 & 42.77 & 47.93 & 56.44 & 23.59 & 51.05 & \textbf{14.55} &  37.43\\
T3A & 29.38 & 27.19 & 16.24 & 58.02  & 40.43 & 30.33 & \underline{46.50} & \underline{53.05} & 60.39 & 26.50 & 49.16 & 13.93 & 37.59\\
\midrule
\OurMethod & \textbf{47.95} & \textbf{43.90} & \textbf{34.73} & \textbf{64.07}  & \textbf{50.08} & \textbf{49.23} & \textbf{52.80} & \textbf{57.69} & \textbf{63.89} & \textbf{46.77} & \textbf{61.88} & \underline{14.24} & \textbf{48.94}\\
\bottomrule
\end{tabular}

\caption{
Top-1 Classification Accuracy (\%) for all corruptions for TANet on three datasets.
\textit{DUA} is evaluated with batch size of 1, following the setting in the original work~\cite{mirza2022dua}. All the other methods are evaluated with batch size of 8. Highest accuracy is shown in bold, while second best is underlined.
}
\label{tab:corr_results_tanet}
\end{table*}

\begin{table*}[!tb]
\centering
\small  
\begin{tabular}{lM{0.75cm}M{0.75cm}M{0.75cm}M{0.75cm}M{0.75cm}M{0.75cm}M{0.75cm}M{0.75cm}M{0.75cm}M{0.75cm}M{0.75cm}M{0.75cm}M{0.75cm}}
\toprule
\multicolumn{14}{c}{Swin UCF101} \\
\midrule
corruptions & gauss & pepper & salt & shot & zoom  & impulse &  defocus & motion &  jpeg & contrast & rain & h265.abr & avg\\
\midrule
Source-Only & 73.01 & 69.57 & 53.32 & 90.38 & 82.66 & 73.83 & 78.19 & 79.21 & 84.03 & 81.58 & 90.69 & 85.33 & 78.48 \\
\midrule
TENT & \underline{77.66} & \underline{75.47} & \underline{60.90} & \underline{91.75} & \underline{85.96} & \underline{78.75} & \textbf{81.29} & \underline{82.86} & \underline{88.13} & \underline{85.81} & \underline{92.20} & \underline{87.42} & \underline{82.35} \\
SHOT & 72.98 & 69.39 & 53.21 & 90.35 & 82.58 & 73.80 & 78.17 & 79.19 & 84.03 & 81.42 & 90.64 & 85.33 &  78.42\\
T3A & 75.89 & 72.56 & 57.34 & 91.22  & 83.87 & 76.79 & \underline{80.02} & 81.28 & 86.52 & 84.17 & 91.59 & 86.89 & 80.68\\
\midrule
\OurMethod & \textbf{81.60} & \textbf{81.52} & \textbf{70.90} & \textbf{92.65}  & \textbf{86.07} & \textbf{82.47} & 79.41 & \textbf{84.55} & \textbf{90.40} & \textbf{86.89} & \textbf{93.21} & \textbf{87.23} & \textbf{84.74}\\
\bottomrule
\end{tabular}

\small  
\begin{tabular}{lM{0.75cm}M{0.75cm}M{0.75cm}M{0.75cm}M{0.75cm}M{0.75cm}M{0.75cm}M{0.75cm}M{0.75cm}M{0.75cm}M{0.75cm}M{0.75cm}M{0.75cm}}
\toprule
\multicolumn{14}{c}{Swin SSv2} \\
\midrule
corruptions & gauss & pepper & salt & shot & zoom  & impulse &  defocus & motion &  jpeg & contrast & rain & h265.abr & avg\\
\midrule
Source-Only & 39.79 & 33.22 & 22.26 & 56.47 & 56.01 & 40.07 & 49.74 & 43.36 & 51.08 & 45.81 & 48.96 & 19.41 & 42.18 \\
\midrule
TENT & 35.83 & \underline{36.61} & 11.92 & \textbf{58.13} & \textbf{57.59} & 40.52 & \underline{53.15} & \underline{46.58} & \underline{53.93} & \underline{51.57} & \underline{52.67} & 15.57 & \underline{42.84} \\
SHOT & \underline{40.42} & 33.82 & \underline{24.07} & 56.31 & 56.01 & \underline{40.67} & 49.59 & 43.79 & 51.31 & 45.53 & 49.46 & 19.66 &  42.55\\
T3A & 39.54 & 33.16 & 22.06 & 56.02  & 55.97 & 39.67 & 50.15 & 44.67 & 52.56 & 45.63 & 49.57 & \underline{19.97} & 42.41\\
\midrule
\OurMethod & \textbf{48.0} & \textbf{43.10} & \textbf{40.09} & \underline{58.04}  & \underline{57.28} & \textbf{48.87} & \textbf{54.20} & \textbf{50.88} & \textbf{58.18} & \textbf{53.17} & \textbf{59.94} & \textbf{24.22} & \textbf{49.66}\\
\bottomrule
\end{tabular}

\small  
\begin{tabular}{lM{0.75cm}M{0.75cm}M{0.75cm}M{0.75cm}M{0.75cm}M{0.75cm}M{0.75cm}M{0.75cm}M{0.75cm}M{0.75cm}M{0.75cm}M{0.75cm}M{0.75cm}}
\toprule
\multicolumn{14}{c}{Swin K400} \\
\midrule
corruptions & gauss & pepper & salt & shot & zoom  & impulse &  defocus & motion &  jpeg & contrast & rain & h265.abr & avg\\
\midrule
Source-Only & 42.91 & 35.52 & 28.64 & 65.37 & 42.15 & 44.43 & 57.56 & 58.92 & 66.09 & 45.84 & 57.12 & 21.56 & 47.17 \\
\midrule
TENT & \underline{43.94} & 35.59 & 26.38 & 65.90 & 43.67 & \underline{45.76} & \textbf{58.60} & 59.65 & 66.64 & \underline{50.04} & 56.94 & 20.94 & 47.84 \\
SHOT & 43.73 & 36.15 & \underline{29.38} & \underline{66.26} & 43.02 & 45.27 & 58.42 & \underline{59.86} & \underline{67.0} & 46.08 & 58.20 & \underline{22.39} &  47.98\\
T3A & 43.55 & \underline{36.49} & 29.25 & 65.73  & \underline{43.93} & 45.04 & \underline{58.55} & 59.78 & 66.96 & 47.43 & \underline{59.17} & \textbf{22.51} & \underline{48.20}\\
\midrule
\OurMethod & \textbf{54.45} & \textbf{51.17} & \textbf{44.15} & \textbf{67.92}  & \textbf{51.99} & \textbf{55.92} & 57.57 & \textbf{60.97} & \textbf{67.68} & \textbf{54.47} & \textbf{65.97} & 22.34 & \textbf{54.55}\\
\bottomrule
\end{tabular}

\caption{
Top-1 Classification Accuracy (\%) for all corruptions for Video Swin Transformer on three datasets.
\textit{NORM} and \textit{DUA} cannot be evaluated on transformer without batch norm layers. All the methods are evaluated with batch size of 8. Highest accuracy is shown in bold, while second best is underlined.
}
\label{tab:corr_results_swin}
\end{table*}

{\small
\bibliographystyle{ieee_fullname}
\bibliography{egbib}

\begin{thebibliography}{10}\itemsep=-1pt

\bibitem{arnab2021vivit}
Anurag Arnab, Mostafa Dehghani, Georg Heigold, Chen Sun, Mario Lu{\v{c}}i{\'c},
  and Cordelia Schmid.
\newblock Vivit: A video vision transformer.
\newblock In {\em ICCV}, pages 6836--6846, 2021.

\bibitem{azimi2022self}
Fatemeh Azimi, Sebastian Palacio, Federico Raue, J{\"o}rn Hees, Luca
  Bertinetto, and Andreas Dengel.
\newblock Self-supervised test-time adaptation on video data.
\newblock In {\em WACV}, pages 3439--3448, 2022.

\bibitem{bertasius2021space}
Gedas Bertasius, Heng Wang, and Lorenzo Torresani.
\newblock Is space-time attention all you need for video understanding?
\newblock In {\em ICML}, volume~2, page~4, 2021.

\bibitem{boudiaf2022parameter}
Malik Boudiaf, Romain Mueller, Ismail~Ben Ayed, and Luca Bertinetto.
\newblock {Parameter-free Online Test-time Adaptation}.
\newblock In {\em CVPR}, 2022.

\bibitem{carreira2017quo}
Joao Carreira and Andrew Zisserman.
\newblock Quo vadis, action recognition? a new model and the kinetics dataset.
\newblock In {\em CVPR}, pages 6299--6308, 2017.

\bibitem{chen2020simple}
Ting Chen, Simon Kornblith, Mohammad Norouzi, and Geoffrey Hinton.
\newblock A simple framework for contrastive learning of visual
  representations.
\newblock In {\em ICML}, pages 1597--1607. PMLR, 2020.

\bibitem{deng2009imagenet}
Jia Deng, Wei Dong, Richard Socher, Li-Jia Li, Kai Li, and Li Fei-Fei.
\newblock Imagenet: A large-scale hierarchical image database.
\newblock In {\em CVPR}, pages 248--255. Ieee, 2009.

\bibitem{fan2021multiscale}
Haoqi Fan, Bo Xiong, Karttikeya Mangalam, Yanghao Li, Zhicheng Yan, Jitendra
  Malik, and Christoph Feichtenhofer.
\newblock Multiscale vision transformers.
\newblock In {\em ICCV}, pages 6824--6835, 2021.

\bibitem{feichtenhofer2020x3d}
Christoph Feichtenhofer.
\newblock X3d: Expanding architectures for efficient video recognition.
\newblock In {\em CVPR}, 2020.

\bibitem{feichtenhofer2019slowfast}
Christoph Feichtenhofer, Haoqi Fan, Jitendra Malik, and Kaiming He.
\newblock Slowfast networks for video recognition.
\newblock In {\em CVPR}, pages 6202--6211, 2019.

\bibitem{gandelsman2022test}
Yossi Gandelsman, Yu Sun, Xinlei Chen, and Alexei~A Efros.
\newblock Test-time training with masked autoencoders.
\newblock In {\em NeurIPS}, 2022.

\bibitem{ganin2016domain}
Yaroslav Ganin, Evgeniya Ustinova, Hana Ajakan, Pascal Germain, Hugo
  Larochelle, Fran{\c{c}}ois Laviolette, Mario Marchand, and Victor Lempitsky.
\newblock Domain-adversarial training of neural networks.
\newblock {\em JMLR}, 17(1):2096--2030, 2016.

\bibitem{gidaris2018unsupervised}
Spyros Gidaris, Praveer Singh, and Nikos Komodakis.
\newblock {Unsupervised Representation Learning by Predicting Image Rotations}.
\newblock In {\em ICLR}, 2018.

\bibitem{goyal2017something}
Raghav Goyal, Samira Ebrahimi~Kahou, Vincent Michalski, Joanna Materzynska,
  Susanne Westphal, Heuna Kim, Valentin Haenel, Ingo Fruend, Peter Yianilos,
  Moritz Mueller-Freitag, et~al.
\newblock The" something something" video database for learning and evaluating
  visual common sense.
\newblock In {\em ICCV}, pages 5842--5850, 2017.

\bibitem{he2022masked}
Kaiming He, Xinlei Chen, Saining Xie, Yanghao Li, Piotr Doll{\'a}r, and Ross
  Girshick.
\newblock {Masked Autoencoders are Scalable Vision Learners}.
\newblock In {\em CVPR}, 2022.

\bibitem{he2016deep}
Kaiming He, Xiangyu Zhang, Shaoqing Ren, and Jian Sun.
\newblock Deep residual learning for image recognition.
\newblock In {\em CVPR}, pages 770--778, 2016.

\bibitem{iwasawa2021test}
Yusuke Iwasawa and Yutaka Matsuo.
\newblock Test-time classifier adjustment module for model-agnostic domain
  generalization.
\newblock {\em NeurIPS}, 34:2427--2440, 2021.

\bibitem{iwasawa2021t3a}
Yusuke Iwasawa and Yutaka Matsuo.
\newblock {Test-Time Classifier Adjustment Module for Model-Agnostic Domain
  Generalization}.
\newblock {\em NIPS}, 2021.

\bibitem{kay2017kinetics}
Will Kay, Joao Carreira, Karen Simonyan, Brian Zhang, Chloe Hillier, Sudheendra
  Vijayanarasimhan, Fabio Viola, Tim Green, Trevor Back, Paul Natsev, et~al.
\newblock The kinetics human action video dataset.
\newblock {\em arXiv preprint arXiv:1705.06950}, 2017.

\bibitem{kingma2014adam}
Diederik~P Kingma and Jimmy Ba.
\newblock Adam: A method for stochastic optimization.
\newblock {\em arXiv preprint arXiv:1412.6980}, 2014.

\bibitem{kojima2022cfa}
Takeshi Kojima, Yutaka Matsuo, and Yusuke Iwasawa.
\newblock Robustifying vision transformer without retraining from scratch by
  test-time class-conditional feature alignment.
\newblock {\em arXiv preprint arXiv:2206.13951}, 2022.

\bibitem{li2020tea}
Yan Li, Bin Ji, Xintian Shi, Jianguo Zhang, Bin Kang, and Limin Wang.
\newblock Tea: Temporal excitation and aggregation for action recognition.
\newblock In {\em CVPR}, pages 909--918, 2020.

\bibitem{liang2020shot}
Jian Liang, Dapeng Hu, and Jiashi Feng.
\newblock Do we really need to access the source data? source hypothesis
  transfer for unsupervised domain adaptation.
\newblock In {\em ICML}, pages 6028--6039. PMLR, 2020.

\bibitem{lin2019tsm}
Ji Lin, Chuang Gan, and Song Han.
\newblock Tsm: Temporal shift module for efficient video understanding.
\newblock In {\em ICCV}, pages 7083--7093, 2019.

\bibitem{liu2021tttpp}
Yuejiang Liu, Parth Kothari, Bastien van Delft, Baptiste Bellot-Gurlet, Taylor
  Mordan, and Alexandre Alahi.
\newblock {TTT++: When Does Self-Supervised Test-Time Training Fail or Thrive?}
\newblock In {\em NeurIPS}, 2021.

\bibitem{liu2021swin}
Ze Liu, Yutong Lin, Yue Cao, Han Hu, Yixuan Wei, Zheng Zhang, Stephen Lin, and
  Baining Guo.
\newblock Swin transformer: Hierarchical vision transformer using shifted
  windows.
\newblock In {\em ICCV}, pages 10012--10022, 2021.

\bibitem{liu2020teinet}
Zhaoyang Liu, Donghao Luo, Yabiao Wang, Limin Wang, Ying Tai, Chengjie Wang,
  Jilin Li, Feiyue Huang, and Tong Lu.
\newblock Teinet: Towards an efficient architecture for video recognition.
\newblock In {\em AAAI}, volume~34, pages 11669--11676, 2020.

\bibitem{liu2022video}
Ze Liu, Jia Ning, Yue Cao, Yixuan Wei, Zheng Zhang, Stephen Lin, and Han Hu.
\newblock Video swin transformer.
\newblock In {\em CVPR}, pages 3202--3211, 2022.

\bibitem{liu2021tam}
Zhaoyang Liu, Limin Wang, Wayne Wu, Chen Qian, and Tong Lu.
\newblock Tam: Temporal adaptive module for video recognition.
\newblock In {\em ICCV}, pages 13708--13718, 2021.

\bibitem{mirza2022dua}
M~Jehanzeb Mirza, Jakub Micorek, Horst Possegger, and Horst Bischof.
\newblock The norm must go on: Dynamic unsupervised domain adaptation by
  normalization.
\newblock In {\em CVPR}, pages 14765--14775, 2022.

\bibitem{nado2020evaluating}
Zachary Nado, Shreyas Padhy, D Sculley, Alexander D'Amour, Balaji
  Lakshminarayanan, and Jasper Snoek.
\newblock Evaluating prediction-time batch normalization for robustness under
  covariate shift.
\newblock {\em arXiv preprint arXiv:2006.10963}, 2020.

\bibitem{neimark2021video}
Daniel Neimark, Omri Bar, Maya Zohar, and Dotan Asselmann.
\newblock Video transformer network.
\newblock In {\em ICCV}, pages 3163--3172, 2021.

\bibitem{patrick2021keeping}
Mandela Patrick, Dylan Campbell, Yuki Asano, Ishan Misra, Florian Metze,
  Christoph Feichtenhofer, Andrea Vedaldi, and Jo{\~a}o~F Henriques.
\newblock Keeping your eye on the ball: Trajectory attention in video
  transformers.
\newblock In {\em NeurIPS}, volume~34, pages 12493--12506, 2021.

\bibitem{qiu2017learning}
Zhaofan Qiu, Ting Yao, and Tao Mei.
\newblock Learning spatio-temporal representation with pseudo-3d residual
  networks.
\newblock In {\em ICCV}, pages 5533--5541, 2017.

\bibitem{schiappa2022large}
Madeline~C. Schiappa, Naman Biyani, Shruti Vyas, Hamid Palangi, Vibhav Vineet,
  and Yogesh Rawat.
\newblock Large-scale robustness analysis of video action recognition models.
\newblock {\em CoRR}, abs/2207.01398, 2022.

\bibitem{schneider2020norm}
Steffen Schneider, Evgenia Rusak, Luisa Eck, Oliver Bringmann, Wieland Brendel,
  and Matthias Bethge.
\newblock Improving robustness against common corruptions by covariate shift
  adaptation.
\newblock In {\em NeurIPS}, pages 11539--11551, 2020.

\bibitem{soomro2012ucf101}
Khurram Soomro, Amir~Roshan Zamir, and Mubarak Shah.
\newblock Ucf101: A dataset of 101 human actions classes from videos in the
  wild.
\newblock {\em arXiv preprint arXiv:1212.0402}, 2012.

\bibitem{sun2017correlation}
Baochen Sun, Jiashi Feng, and Kate Saenko.
\newblock Correlation alignment for unsupervised domain adaptation.
\newblock In {\em Domain Adaptation in Computer Vision Applications}, pages
  153--171. Springer, 2017.

\bibitem{sun2016coral}
Baochen Sun and Kate Saenko.
\newblock {Deep CORAL: Correlation Alignment for Deep Domain Adaptation}.
\newblock 2016.

\bibitem{sun2020ttt}
Yu Sun, Xiaolong Wang, Zhuang Liu, John Miller, Alexei Efros, and Moritz Hardt.
\newblock {Test-Time Training with Self-Supervision for Generalization under
  Distribution Shifts}.
\newblock 2020.

\bibitem{tran2015learning}
Du Tran, Lubomir Bourdev, Rob Fergus, Lorenzo Torresani, and Manohar Paluri.
\newblock Learning spatiotemporal features with 3d convolutional networks.
\newblock In {\em CVPR}, pages 4489--4497, 2015.

\bibitem{truong2022direcformer}
Thanh-Dat Truong, Quoc-Huy Bui, Chi~Nhan Duong, Han-Seok Seo, Son~Lam Phung,
  Xin Li, and Khoa Luu.
\newblock Direcformer: A directed attention in transformer approach to robust
  action recognition.
\newblock In {\em CVPR}, pages 20030--20040, 2022.

\bibitem{van2008visualizing}
Laurens Van~der Maaten and Geoffrey Hinton.
\newblock Visualizing data using t-sne.
\newblock {\em JMLR}, 9(11), 2008.

\bibitem{wang2021tent}
Dequan Wang, Evan Shelhamer, Shaoteng Liu, Bruno Olshausen, and Trevor Darrell.
\newblock Tent: Fully test-time adaptation by entropy minimization.
\newblock In {\em ICLR}, 2021.

\bibitem{wang2021tdn}
Limin Wang, Zhan Tong, Bin Ji, and Gangshan Wu.
\newblock Tdn: Temporal difference networks for efficient action recognition.
\newblock In {\em CVPR}, pages 1895--1904, 2021.

\bibitem{wang2016temporal}
Limin Wang, Yuanjun Xiong, Zhe Wang, Yu Qiao, Dahua Lin, Xiaoou Tang, and Luc
  Van~Gool.
\newblock Temporal segment networks: Towards good practices for deep action
  recognition.
\newblock In {\em ECCV}, pages 20--36. Springer, 2016.

\bibitem{wang2022continual}
Qin Wang, Olga Fink, Luc Van~Gool, and Dengxin Dai.
\newblock Continual test-time domain adaptation.
\newblock In {\em CVPR}, pages 7201--7211, 2022.

\bibitem{wang2021action}
Zhengwei Wang, Qi She, and Aljosa Smolic.
\newblock Action-net: Multipath excitation for action recognition.
\newblock In {\em CVPR}, 2021.

\bibitem{wu2021rethinking}
Yuxin Wu and Justin Johnson.
\newblock Rethinking" batch" in batchnorm.
\newblock {\em arXiv preprint arXiv:2105.07576}, 2021.

\bibitem{xiang2022spatiotemporal}
Wangmeng Xiang, Chao Li, Biao Wang, Xihan Wei, Xian-Sheng Hua, and Lei Zhang.
\newblock Spatiotemporal self-attention modeling with temporal patch shift for
  action recognition.
\newblock In {\em ECCV}, 2022.

\bibitem{yang2020temporal}
Ceyuan Yang, Yinghao Xu, Jianping Shi, Bo Dai, and Bolei Zhou.
\newblock Temporal pyramid network for action recognition.
\newblock In {\em CVPR}, 2020.

\bibitem{yang2022recurring}
Jiewen Yang, Xingbo Dong, Liujun Liu, Chao Zhang, Jiajun Shen, and Dahai Yu.
\newblock Recurring the transformer for video action recognition.
\newblock In {\em CVPR}, pages 14063--14073, 2022.

\bibitem{yi2021benchmarking}
Chenyu Yi, Siyuan Yang, Haoliang Li, Yap{-}Peng Tan, and Alex~C. Kot.
\newblock Benchmarking the robustness of spatial-temporal models against
  corruptions.
\newblock In {\em NeurIPS}, 2021.

\bibitem{yin2020dreaming}
Hongxu Yin, Pavlo Molchanov, Jose~M Alvarez, Zhizhong Li, Arun Mallya, Derek
  Hoiem, Niraj~K Jha, and Jan Kautz.
\newblock Dreaming to distill: Data-free knowledge transfer via deepinversion.
\newblock In {\em CVPR}, pages 8715--8724, 2020.

\bibitem{zellinger2017central}
Werner Zellinger, Thomas Grubinger, Edwin Lughofer, Thomas Natschl{\"a}ger, and
  Susanne Saminger-Platz.
\newblock Central moment discrepancy (cmd) for domain-invariant representation
  learning.
\newblock {\em arXiv preprint arXiv:1702.08811}, 2017.

\bibitem{zhang2021token}
Hao Zhang, Yanbin Hao, and Chong-Wah Ngo.
\newblock Token shift transformer for video classification.
\newblock In {\em ACM Multimedia}, pages 917--925, 2021.

\bibitem{zhang2021memo}
Marvin Zhang, Sergey Levine, and Chelsea Finn.
\newblock Memo: Test time robustness via adaptation and augmentation.
\newblock {\em arXiv preprint arXiv:2110.09506}, 2021.

\end{thebibliography}
}

\end{document}